%% file: DnC.tex
\documentclass[final]{cvpr}

\usepackage{stackengine}
\usepackage{times}
\usepackage{epsfig}
\usepackage{graphicx}
\usepackage{amsmath}
\usepackage{amssymb}
\usepackage{tabularx}
\usepackage{booktabs}
\usepackage{multirow}
\usepackage{mathtools}
\usepackage{mathabx}
\usepackage{rotating}
\usepackage{dashrule}
\usepackage[usernames,dvipsnames,svgnames,table]{xcolor}
\usepackage[super]{nth}

\usepackage{color, colortbl}

\definecolor{pearDark}{HTML}{2980B9}
\definecolor{DnCBG}{rgb}{0.9, 0.9, 1.}

\definecolor{Gray}{gray}{0.5}
\definecolor{GrayBG}{gray}{0.95}
\definecolor{BlueBG}{rgb}{0.9, 0.9, 1.}

\definecolor{Gray}{gray}{0.5}
\definecolor{pearDark}{HTML}{2980B9}
\definecolor{nicergreen}{rgb}{0.13, 0.54, 0.13}
\definecolor{nicered}{rgb}{0.83, 0.16, 0.16}
\newcommand\showdiff[1]{\textbf{\textcolor{nicergreen}{#1}}}
\newcommand\showdiffn[1]{\textbf{\textcolor{nicered}{#1}}}

\newcommand*{\mathcolor}{}
\def\mathcolor#1#{\mathcoloraux{#1}}
\newcommand*{\mathcoloraux}[3]{%
  \protect\leavevmode
  \begingroup
    \color#1{#2}#3%
  \endgroup
}

\newcommand{\cgaphl}[2]{
\fontsize{6pt}{1em}\selectfont{\textcolor{nicergreen}{(${#1}$\textbf{#2})}}
}

\newcolumntype{C}{>{\centering\arraybackslash}X}

\def\delequal{\mathrel{\ensurestackMath{\stackon[1pt]{=}{\scriptstyle\Delta}}}}

\definecolor{customcolor}{rgb}{0.82, 0.41, 0.12}

\newcommand{\normtwo}[1]{\left\|#1\right\|_2}
\newcommand\baseline{MoCLR}
\newcommand\jft{JFT-300M}

\newcommand{\apbbox}[1]{AP$^\text{bb}_\text{#1}$}
\newcommand{\apmask}[1]{AP$^\text{mk}_\text{#1}$}

\newcommand{\bOne}{{\bf 1}}
\newcommand{\pa}[1]{\left(#1\right)}

\newcommand\rot[1]{\begin{rotate}{45}#1 \end{rotate} }

\usepackage[pagebackref=true,breaklinks=true,letterpaper=true,colorlinks,bookmarks=false]{hyperref}

\def\cvprPaperID{5539} %
\def\confYear{CVPR 2021}

\begin{document}

\title{Divide and Contrast: Self-supervised Learning from Uncurated Data}

\author{Yonglong Tian
\textsuperscript{*} 
\\
MIT
\and
Olivier J. Hénaff \\
DeepMind
\and
Aaron van den Oord \\ 
DeepMind
}

\maketitle
\let\thefootnote\relax\footnotetext{*Work done while interning at DeepMind.
}

\input{tex/abstract}

\input{tex/introduction}
\input{tex/related_work}

\input{tex/method}
\input{tex/experiments_uncurated}
\input{tex/experiments_imagenet}
\input{tex/conclusion}

{\small
\bibliographystyle{ieee_fullname}
\bibliography{DnC}
}

\clearpage
\appendix

\input{tex/sup.tex}

\end{document}

%% file: tex/abstract.tex
\begin{abstract}

Self-supervised learning holds promise in leveraging large amounts of unlabeled data, however much of its progress has thus far been limited to highly curated pre-training data such as ImageNet. We explore the effects of contrastive learning from larger, less-curated image datasets such as YFCC, and find there is indeed a large difference in the resulting representation quality. We hypothesize that this curation gap is due to a shift in the distribution of image classes---which is more diverse and heavy-tailed---resulting in less relevant negative samples to learn from. We test this hypothesis with a new approach, Divide and Contrast (DnC), which alternates between contrastive learning and clustering-based hard negative mining. When pretrained on less curated datasets, DnC greatly improves the performance of self-supervised learning on downstream tasks, while remaining competitive with the current state-of-the-art on curated datasets.

\end{abstract}

%% file: tex/introduction.tex
\section{Introduction}

Recent developments in self-supervised learning have shown that it is possible to learn high-level representations of object categories from unlabeled images~\cite{he2020momentum, henaff2019data,tian2019contrastive,chen2020simple,wu2018unsupervised}, phonetic information from speech \cite{oord2018representation, schneider2019wav2vec} and language understanding from raw text \cite{devlin2018bert,yang2019xlnet}. The most studied benchmark in self-supervised learning is ImageNet~\cite{deng2009imagenet}, where representations learned from unlabeled images can surpass supervised representations, both in terms of their data-efficiency and transfer-learning performance \cite{chen2020big, grill2020bootstrap}.

\begin{figure}[t]
  \centering
  \includegraphics[width=0.98\linewidth]{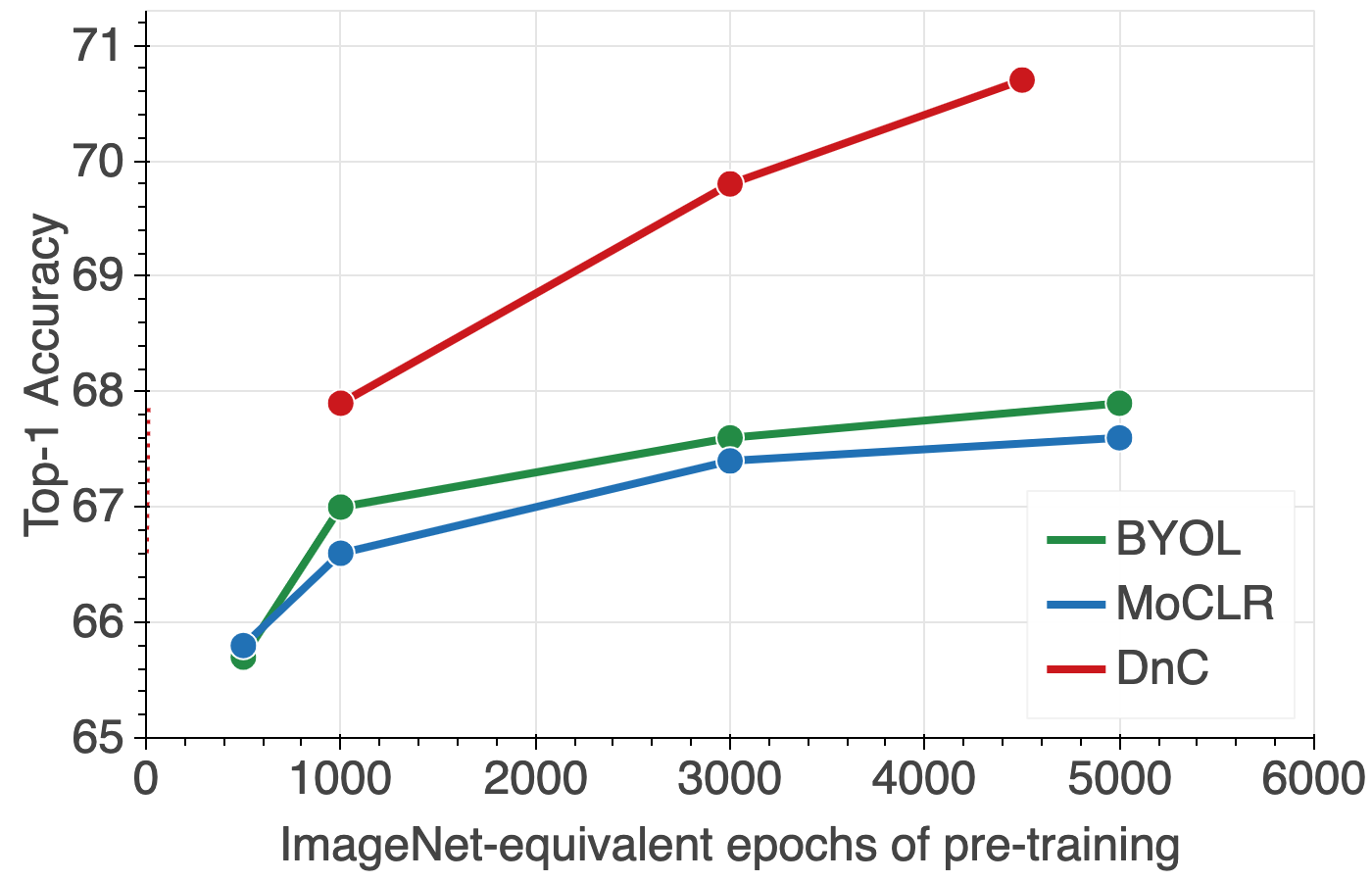}
  \caption{Linear evaluation on ImageNet of representations learned from a large-scale uncurated dataset using ResNet-50. Divide and Contrast (DnC) is better able to handle the diverse and long-tailed distribution of images and improves more with longer training. X-axis represents total computation, in ImageNet-equivalent epochs. %
  }\label{fig:teaser}
\end{figure}

One of the caveats with self-supervised learning on ImageNet is that it is not completely ``self-supervised". 
The training set of ImageNet, on which the representations are learned, is heavily curated and required extensive human effort to create \cite{deng2009imagenet}. 
In particular, ImageNet contains many fine-grained classes (such as subtly different dog breeds), each one containing roughly the same number of images. 
While this \textit{consistency} may facilitate the learning of high-level visual representations, limiting self-supervised learning to such curated datasets risks biasing their development towards methods which require this consistency, limiting their applicability to more diverse downstream tasks and larger datasets for pre-training.

In this paper we assess how well recent self-supervised learning methods perform on downstream tasks (including ImageNet) when they are pre-trained on significantly less curated datasets, such as YFCC100M \cite{thomee2016yfcc100m}. We observe a notable drop in performance of over 9\% Top-1 accuracy (from 74.3\% to 65.3\%) for a ResNet50 model trained with the current state-of-the-art in self-supervised learning.

We hypothesize that this \emph{curation gap} is due to the heavy-tailed nature of images collected in the wild, which present much more diverse content, breaking the \textit{global consistency} exploited in previous datasets. We test this hypothesis with a new method, Divide and Contast (DnC), which attempts to recover
\textit{local consistency} in subsets of the larger, uncurated dataset, such that self-supervised learning methods can learn high-level features that are specific to each subset. We find that such semantically coherent subsets can be straightforwardly obtained by clustering the representations of standard self-supervised models.

Divide and Contrast (DnC) proceeds by training individual ``expert'' models on each subset and distilling them into a single model. As a result, DnC can be used in combination with any self-supervised learning technique, and requires the same amount of computation, as each expert is trained for significantly less time. Finally, this computation is trivially parallelized, allowing it to be scaled to massive datasets. %

The remainder of this paper is structured as follows. We first review related work in self-supervised learning. We then present a new stronger baseline (\baseline{}) which improves over current contrastive methods, matching the performance of the current state-of-the-art (BYOL  \cite{grill2020bootstrap}). Next we present the main method, Divide and Contrast, and how this model can be used together with any SSL method. In the experiments we evaluate the different hypotheses that support DnC, and compare its ability to learn from uncurated dataset with existing methods.

%% file: tex/related_work.tex
\section{Related Work}

Recent self-supervised representation learning generally includes three types of methods: \emph{generative models} that directly model the data distribution, \emph{pretext tasks} that are manually designed according to the data, and \emph{contrastive learning} that contrasts positive pairs with negative pairs.

\noindent\textbf{Generative models.} While the primary goal of generative models such as GAN~\cite{goodfellow2014generative,chen2016infogan} or VAE~\cite{kingma2013auto} is to model the data distribution (e.g., sample new data or estimate likelihood), the encoder network can also extract good representations~\cite{radford2015unsupervised}. Recent state of the art generative models for representation learning include BiGAN~\cite{donahue2016adversarial} and BigBiGAN~\cite{donahue2019large}, which learn a bidirectional mapping between the latent codes and the images, and iGPT~\cite{chen2020generative} which trains an autoregressive model on raw pixels.

\noindent\textbf{Pretext tasks.} Good representations may also be learned by solving various pretext tasks. Examples include denoising~\cite{vincent2008extracting}, relative patch prediction~\cite{doersch2015unsupervised}, image inpainting~\cite{pathak2016context}, noise prediction~\cite{bojanowski2017unsupervised}, colorization~\cite{zhang2016colorful,zhang2017split,vondrick2018tracking}, Jigsaw~\cite{noroozi2016unsupervised}, exemplar modeling~\cite{dosovitskiy2014discriminative}, motion segmentation~\cite{pathak2017learning}, image transformation prediction~\cite{gidaris2018unsupervised,zhang2019aet}, tracking~\cite{wang2015unsupervised}, or even the combination of multiple tasks~\cite{doersch2017multi}. Another line of methods generates pseudo labels by clustering features~\cite{caron2018deep,caron2019unsupervised,ji2019invariant,zhan2020online,alwassel2019self}. Most recently, SeLa~\cite{Asano2020} jointly clusters images and balances the clusters. SwAV~\cite{caron2020unsupervised} learns representatons by having different views of the same image assigned to the same cluster. Another work~\cite{hsu2018unsupervised} directly optimizes the transferability of representation by integrating clustering with meta-learning.

\noindent\textbf{Contrastive learning.} Contrastive learning is a widely-used generic method. The loss function for contrastive learning has evolved from early margin-based binary classification~\cite{hadsell2006dimensionality}, to triplet loss~\cite{schroff2015facenet}, and to recent k-pair loss~\cite{sohn2016improved,oord2018representation}. The core idea lying at the heart of the recent series of self-supervised contrastive learning methods~\cite{wu2018unsupervised,oord2018representation,henaff2019data,tian2019contrastive,zhuang2019local,bachman2019learning,he2020momentum,misra2019self,chen2020simple,chen2020improved,tian2020makes,chen2020big,li2020prototypical,cao2020parametric} is to maximize the agreement between two ``views'' of the same image while repulsing ``views'' from different images. Such views can be created by color decomposition~\cite{tian2019contrastive}, patch cropping~\cite{oord2018representation,henaff2019data,bachman2019learning}, data augmentation~\cite{chen2020simple,chen2020big,srinivas2020curl}, or image segmentation \cite{henaff2021efficient, van2021unsupervised, zhang2020self}. Indeed, contrastive learning is very general such that it can be easily adapted to different data types. Examples include different frames of video~\cite{oord2018representation,zhuang2019unsupervised,sermanet2018time,han2019video,gordon2020watching,han2020memory}, point clouds~\cite{xie2020pointcontrast}, multiple sensory data~\cite{morgado2020audio,chung2019perfect,patrick2020multi}, text and its context~\cite{mikolov2013distributed,yang2019xlnet,logeswaran2018efficient,Kong2020A}, or video and language~\cite{sun2019contrastive,miech2019end,li2020learning}. A set of other work~\cite{arora2019theoretical,tian2020makes,zhao2020makes,xiao2020should,tosh2020contrastive,purushwalkam2020demystifying,wang2020understanding} focuses on providing empirical and theoretical understanding of contrastive learning. Recently a non-contrastive method BYOL~\cite{grill2020bootstrap} applies a momentum-encoder to one view and predicts its output from the other, inspired by bootstrapping RL~\cite{guo2020bootstrap}. Finally, contrastive learning has also been applied to supervised image classification~\cite{khosla2020supervised}, image translation~\cite{park2020contrastive}, knowledge distillation~\cite{Tian2020Contrastive,rao2020unified}, and adversarial learning~\cite{kim2020adversarial}.

This paper is also related to knowledge distillation~\cite{hinton2015distilling}. In~\cite{hinton2015distilling}, several expert models were also trained in parallel on a large scale dataset, and then distilled into a single model. While labels are assumed available in~\cite{hinton2015distilling} to partition the dataset and distill into a single model, we are dealing with self-supervised learning without supervision. Our distillation procedure is also inspired by FitNet~\cite{romero2014fitnets}. 

Lastly, while self-supervised representation learning on uncurated datasets is largely unexplored, there are a few prior attempts~\cite{caron2019unsupervised,goyal2019scaling}. In~\cite{caron2019unsupervised}, clustering is applied to generate training targets, and in order to capture the long-tailed distribution of images in the uncurated YFCC100m~\cite{thomee2016yfcc100m}, a hierachical formulation is proposed. The work of~\cite{goyal2019scaling} benchmarked pretext-based self-supervised methods in a large scale setting, e.g., jigSaw, colorization and rotation prediction, and found that these pretext tasks are not `hard' enough to take full advantage of large scale data. Concurrent work SEER~\cite{goyal2021self} directly scales up SwAV with larger models and datasets.

%% file: tex/method.tex
\section{Divide and Contrast}

\begin{figure*}[t]
  \centering
  \includegraphics[width=0.97\textwidth]{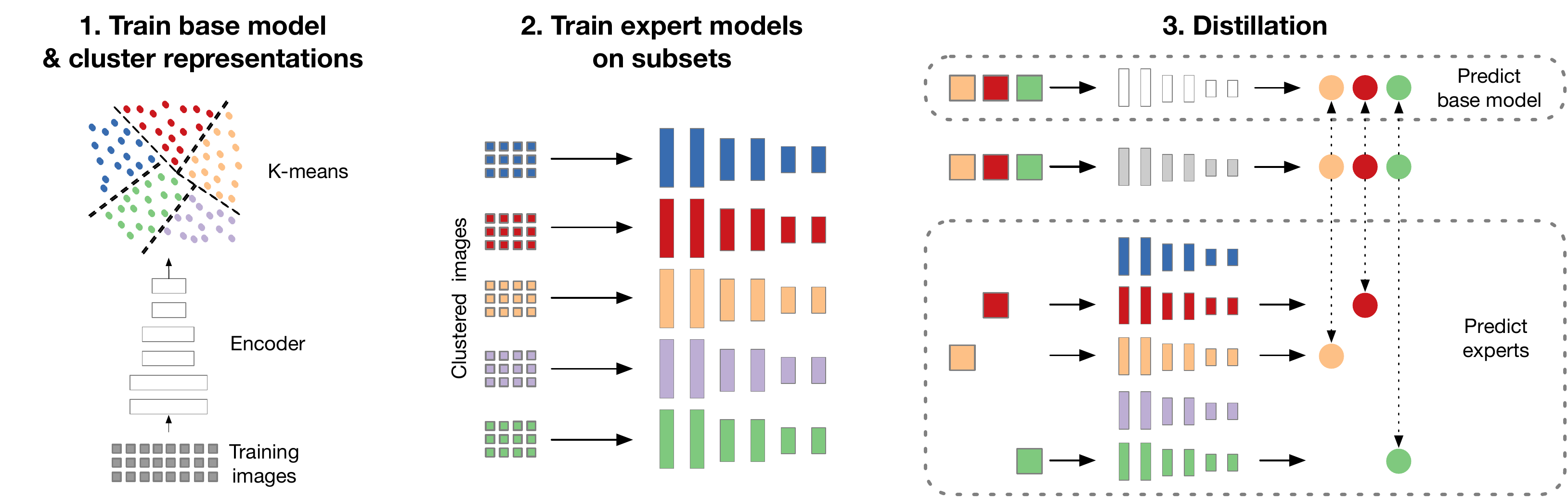}
  \caption{Overview of Divide and Contrast (DnC). DnC can be used in conjunction with any self-supervised learning method (we use \baseline{}, an improvement to SimCLR). In the first step a self-supervised learning method is trained on the whole dataset, which we call the base model. The image representations of the base model are then clustered with k-means into 5, 10 or more groups. In the second step, the clustered dataset is then used to train an expert model on each of the image clusters. In the third step the experts and base model are distilled into a single model by predicting their representations. By splitting the dataset into semantically-similar subsets, contrastive methods need to pay more attention to the differences between the images in those clusters and learn more specific representations.}\label{fig:dnc_model}
\end{figure*}

Though Divide and Contrast can be used in combination with any self-supervised learning technique, in this paper we will combine it with recent state of the art techniques (BYOL, SimCLR, MoCo), such that the model can be compared to a strong baseline and make the experiments relevant with respect to recent developments in the literature. We will start by introducing our baseline, \baseline{}, which is a simple hybrid based on BYOL~\cite{grill2020bootstrap}, SimCLR~\cite{chen2020simple} and MoCo~\cite{he2020momentum}, and as a contrastive method outperforms SimCLR v2~\cite{chen2020big}, achieving similar performance to BYOL (by using a momentum encoder similar to BYOL and MoCo). Even though DnC can be coupled with BYOL, empirically we have found it to work better with methods that use a contrastive loss.

\subsection{An Improved Contrastive Baseline: \baseline{}}
\label{section:baseline}

\baseline{} roughly uses a similar setup to SimCLR and BYOL, we will briefly describe the main components of this setup and highlight the differences.

\noindent\textbf{Augmenting two views.} %
Given an image $x$ and two distributions of image augmentations $\mathcal{T}$ and $\mathcal{T}'$, two \emph{views} are created $v\delequal t(x)$ and $v'\delequal t'(x)$ by respectively applying random image augmentations $t\sim \mathcal{T}$ and $t'\sim \mathcal{T'}$ from these distributions. 
The augmentations $\mathcal{T}$ and $\mathcal{T}'$ here are exactly the same as in BYOL~\cite{grill2020bootstrap}.

\noindent\textbf{Architecture.}
The first augmented view $v$ is fed into an online encoder $f(\cdot)$, followed up with an MLP projection head $g(\cdot)$ to produce a projection $z$. Similarly, an exponential moving average of the online encoder and projection head, also known as momentum encoder~\cite{he2020momentum} or mean teacher~\cite{tarvainen2017mean}, is applied on the second view $v'$ to generate $z'$. The MLP head consists of two layers with a hidden dimension of 4096 and output size of 256, similar to BYOL~\cite{grill2020bootstrap}. 

\noindent\textbf{Loss function.} Given a batch $\mathcal{B}$, we follow the InfoNCE loss \cite{oord2018representation} with a cosine similarity function $s(\cdot,\cdot)$ and a scalar temperature value $\tau$:
\begin{equation}\label{eq:infonce_loss}
    \mathcal{L}_\text{NCE} = -\frac{1}{|\mathcal{B}|}\sum_{i \in \mathcal{B}}\log \frac{e^{s(z_i, z'_i)/\tau}}{e^{s(z_i, z'_i)/\tau} + \sum_{j \in \mathcal{B}/i}e^{s(z_i, z'_j)/\tau}}
\end{equation}
We symmetrize the loss $\mathcal{L}_\text{NCE}$ by separately feeding $v'$ to the online network and $v$ to the momentum encoder, resulting in $\widetilde{\mathcal{L}}_\text{NCE}$. 
The final loss is $\mathcal{L} = \mathcal{L}_\text{NCE} + \widetilde{\mathcal{L}}_\text{NCE}$.

\begin{table}[t]
\caption{Evaluating the \baseline{} baseline which is used in the proposed Divide And Contrast (DnC) model. This comparison is on the ImageNet linear classification benchmark. \baseline{} is a contrastive method which achieves similar performance as BYOL, while requiring only a small change to SimCLR (see Section \ref{section:baseline}).}
\label{tab:linear_baseline}
\begin{center}
\begin{small}
\begin{tabular}{lcccc}
\toprule
Method & Epochs & Top-1 & Top-5 \\
\midrule 
SimCLR \cite{chen2020simple} & 1000 & 69.3 & 89.0 \\
SimCLR v2 \cite{chen2020big} & 1000 & 71.7 & 90.4 \\
MoCo v3 \cite{chen2021empirical} & 800 & 73.8 & - \\
BYOL \cite{grill2020bootstrap} & 1000 & 74.3 & 91.6 \\
\baseline{} (ours) & 1000 & 74.3 & 92.2  \\ 
\bottomrule
\end{tabular}
\end{small}
\end{center}
\end{table}

The difference between \baseline{} and other standard methods are as follows. Compared with SimCLR~\cite{chen2020simple}, we use a momentum encoder, and double the size of the projection head (from 2048 to 4096 for the hidden layer, and from 128 to 256 for the output layer). In comparison with BYOL~\cite{grill2020bootstrap}, we remove the predictor head and use the contrastive loss instead of the mean squared prediction loss. While concurrent work MoCo v3~\cite{chen2021empirical} inherits from BYOL the asymmetric ``projector \& predictor`` design (the online encoder has an additional predictor compared to the momentum network), our MoCLR removes the predictor for simplicity.

In our experiments we set the batch size to 4096 and do not use a memory buffer~\cite{wu2018unsupervised,tian2019contrastive,he2020momentum}. As we will show in our experiments, with these simple changes our \emph{\baseline{}} baseline trained for 1,000 epochs outperforms SimCLR v1/v2~\cite{chen2020big} and is on par with BYOL~\cite{grill2020bootstrap}, see Table~\ref{tab:linear_baseline}. %

\subsection{Divide and Contrast}

The motivation behind Divide and Contrast is that, when training on diverse, large-scale datasets, the density of informative negatives will be sparse if we sample randomly from the whole dataset. Instead, if we contrast locally between semantically-similar classes, the sampled negatives will be more informative and the learned model will capture a more discriminative representation.

As visualized in Figure \ref{fig:dnc_model}, the training of our DnC model consists of three stages:

(1) We first train a \baseline{} model on the given dataset for $N_1$ epochs (though other self-supervised learning methods can be used as well). We will call it the \textbf{base} model. %
We use the base model to extract representations for a set of samples in the training set, and cluster them in to $K$ clusters. With these clusters we partition the dataset into $K$ subsets. 

(2) For each subset, we train a separate \baseline{} model from scratch, which we call \textbf{expert} models. In this stage, we distribute a \emph{total} computational budget of $N_2$ epochs (measured on the whole dataset) to these expert models, proportionally to their corresponding cluster sizes.

(3) Finally, given a base model that captures the general knowledge of the dataset and expert models focusing on locally similar categories, we distill knowledge from these models into a \textbf{distillation} model. In this stage, we train for $N_3$ epochs.

The encoder-architecture for the base model, expert models, and the distillation model are all identical. Therefore, the computational footprint can roughly be measured by summing up the training epochs across all three stages, resulting in a total training of $N_1+N_2+N_3$ epochs (except for the clustering overhead and extra forward pass during distillation, which we discuss later).

\subsection{Distillation}
\label{section:distill}

\begin{figure}[t]
  \centering
  \includegraphics[trim=35 0 0 0, clip, width=0.7\linewidth]{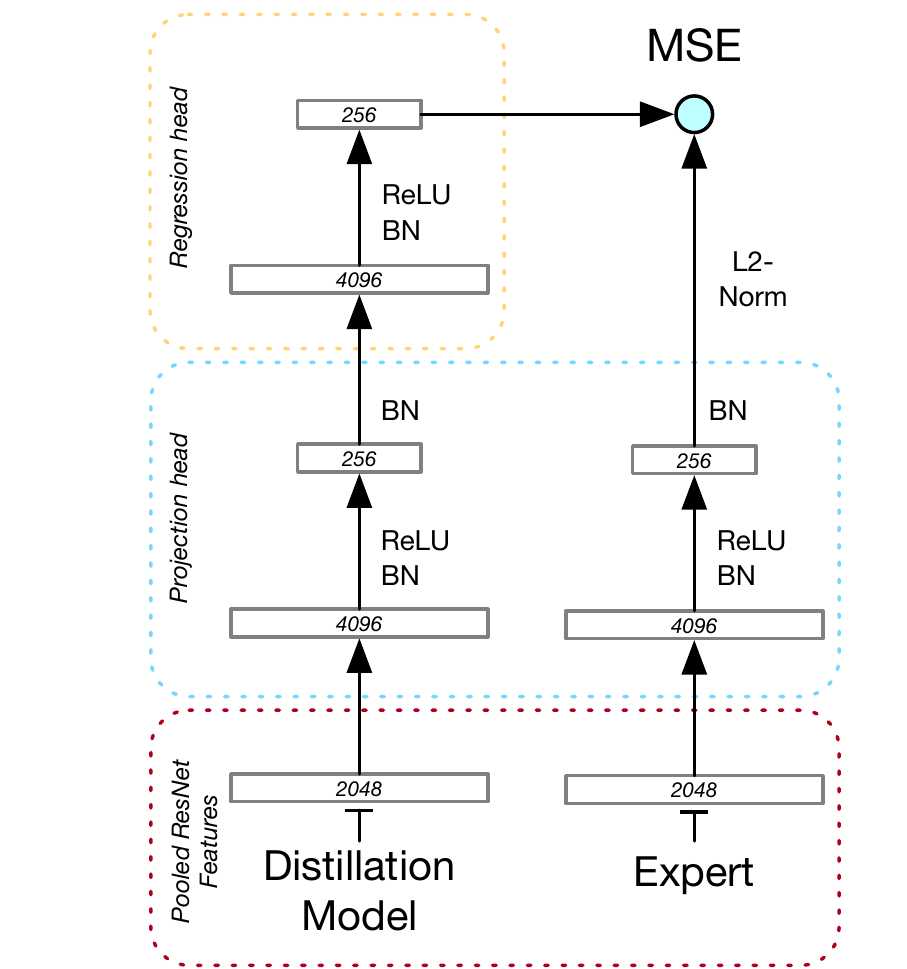}
  \caption{Detailed architectural overview of the distillation of the expert's features. The distillation model and experts are applied to the same augmented image view.}\label{fig:distillation}
\end{figure}

To leverage the information learned by each of the different experts and more general information from the base model, we distill their representation into one model in the last stage of training. During the distillation we use a single augmented image (instead of the 2-view setup) in combination with a simple regression loss to predict the representations in these models (no contrastive loss).

The distillation model's architecture is mostly identical to the other models and is visualized in Figure \ref{fig:distillation}. All have a backbone encoder $f(\cdot)$ and an MLP projection head $g(\cdot)$. %
On top of the projection head in the distillation model there are $K + 1$ regression networks: $r_k(\cdot)$, $k=1,...,K$, one to predict each of the $K$ expert models and another regression network $r_\text{b}(\cdot)$ to predict the base model. The architecture of these regressors is the same as the projecton head, except that we remove the final global BatchNorm after the last output layer.

For distillation we use the same augmentation as during self-supervised learning. Given an augmented input image $x$ with clustering id $k$, we feed it into the distillation model to produce the projection-head output $z$. Similarly we get $z_{b}$ and $z_{k}$ from the base model and the $k$-th expert model respectively. We also $\ell_2$-normalize $z_{b}$ and $z_{k}$ to be unit-norm. %
The distillation objective is then the average of the two mean squared errors:
\vspace{-3pt}
\begin{equation}\label{eq:distill_loss}
\mathcal{L}(x) = \frac{1}{2}\normtwo{r_\text{b}(z) - z_b}^2 + \frac{1}{2}\normtwo{r_k(z) - z_k}^2 \\
\vspace{-3pt}
\end{equation}
Note that the outputs of $r_\text{b}$ and $r_\text{k}$ are \emph{not} $\ell_2$ normalized. 

To make it possible to compare to our baseline methods in terms of the number of epochs trained, we use two augmented views from the same input image and average their losses. This is otherwise not necessary and alternatively one could also increase the batch-size.

The computational cost in this stage is slightly higher than the self-supervised learning stage (e.g., BYOL and \baseline{}), as there are now \textbf{two} forward passes (for the expert and base model) for each view (not backward pass and gradient computation). In contrast, BYOL and \baseline{} only need \textbf{one} forward pass from the momentum encoder. However, we found that always feeding a center crop to expert and base models only leads to very marginal drop in performance (instead of an augmented view). This strategy offers the possibility of first doing a single forward pass over the dataset and storing the activations offline.

%% file: tex/experiments_uncurated.tex
\section{Experiments}

In this section, we compare DnC to BYOL and \baseline{} by pre-training on two large-scale uncurated datasets and evaluating transfer performance on different downstream tasks.

\begin{figure}[t]
  \centering
  \includegraphics[width=0.96\linewidth]{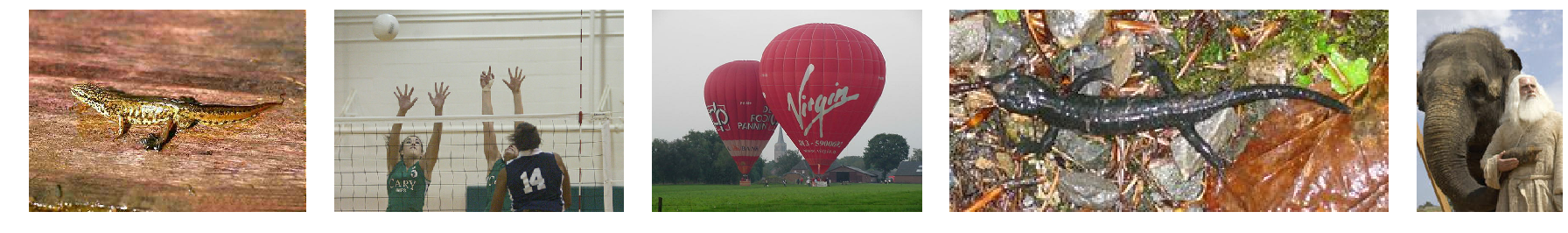}
  \includegraphics[width=0.96\linewidth]{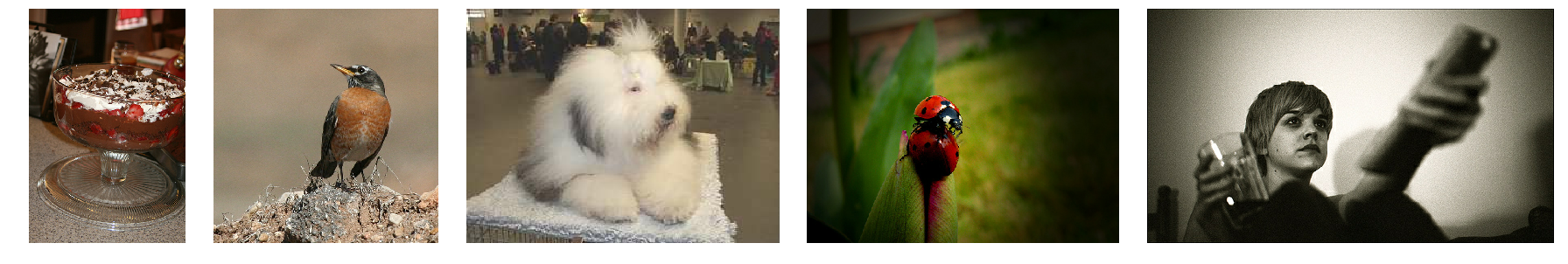}
  \caption{\label{fig:datasets_imagenet} Example ImageNet images}
  \includegraphics[width=0.96\linewidth]{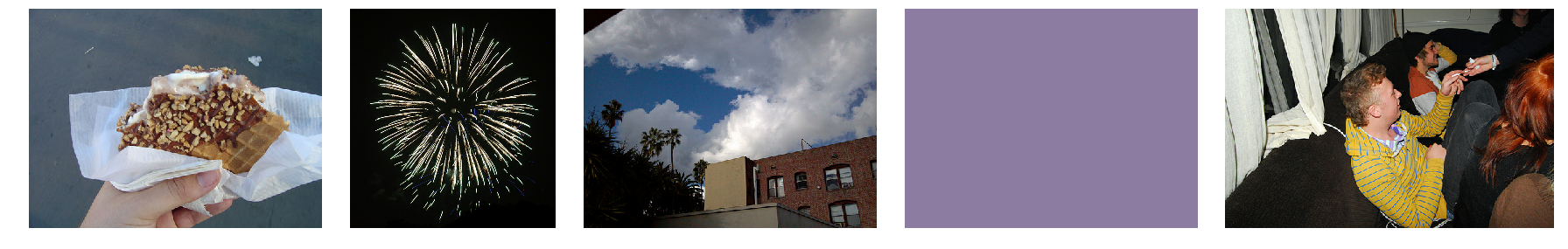}
  \includegraphics[width=0.96\linewidth]{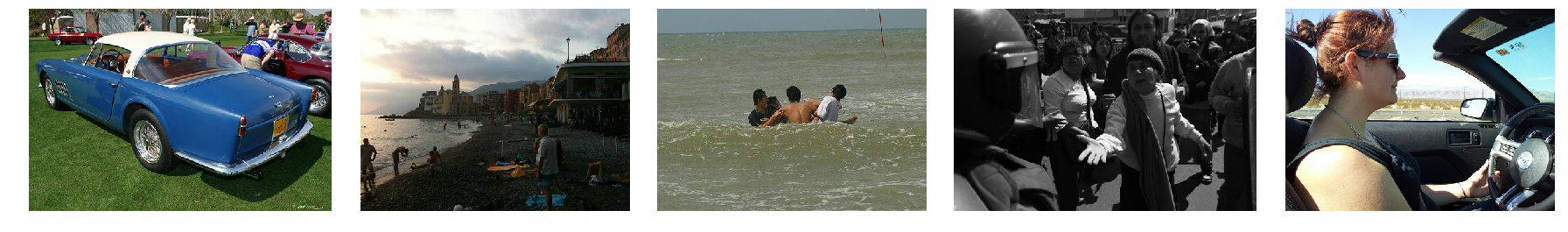}
  \caption{\label{fig:datasets_yfcc} Example YFCC100M images}
\end{figure}

\noindent \textbf{Datasets.} We consider two large-scale uncurated  datasets. The first is a private dataset of roughly 300 million images (\jft{}~\cite{sun2017revisiting}). For the second dataset we use YFCC100M \cite{thomee2016yfcc100m}, a public dataset of 95M Flickr images available under the Creative Commons license. Figures \ref{fig:datasets_imagenet} and \ref{fig:datasets_yfcc} show a visual comparison between images from ImageNet and YFCC100M. ImageNet images often contain the object or animal of interest in the center of the image. ImageNet also does not have a long-tailed distribution (e.g., power law) over object-classes but only considers a specific set of 1000 different classes, which are (roughly) equally represented in the dataset. As a result, specific objects or animals (e.g., common tench, Bedlington terrier, \dots) are over-represented compared to more typically occurring scenes such as human faces and landscapes (which are better represented in YFCC100M). 

\noindent \textbf{Settings.} ResNet-50~\cite{he2016deep} is used in all experiments, unless noted otherwise. For ease of comparison, we report the computational footprint of all experiments in ImageNet-epoch equivalents %
(\eg 1 ``epoch'' $=$ $1281167/\text{batch\textunderscore size}$ iterations). More implementation and optimization details are included in Appendix.

\begin{table}[ht]
\caption{We consider three different training schedules for DnC. The number of total training epochs in each stage, as well as the number of clusters, is specified.}

\label{tab:dnc_schedule}
\begin{center}
\begin{small}
\begin{tabular}{l|c|c|c}
\toprule
\multirow{2}{*}{Schedules} & Base & Experts & Distillation \\
& epochs & total epochs & epochs \\
\hline
1,000 epochs & \hspace{1ex} 200 & \hspace{1ex} 600 (5 clusters) & 200 \\
3,000 epochs & 1,000 & 1,500 (5 clusters) & 500 \\
4,500 epochs & 1,000 & \hspace{1ex}3,000 (10 clusters) & 500 \\
\bottomrule
\end{tabular}
\end{small}
\end{center}
\vspace{-10pt}
\end{table}

\noindent \textbf{DnC Schedules.} 
Table~\ref{tab:dnc_schedule} shows three training schedules with different number of epochs. For example, in the schedule of 3,000 epochs, we first train the base model for 1,000 epochs, after which we cluster the samples into 5 groups. The 5 experts are trained in parallel on these subsets. We use 1,500 epochs in total, spread out over the experts according to the number of images in each cluster (300 on average per expert). The distillation model is then trained for 500 epochs. 
See Section~\ref{sec:run_time} for analysis of run time.

\subsection{Linear Evaluation on ImageNet and Places-365}

\input{fig_tex/table_uncurated_v4}

Table \ref{tab:uncurated_results} shows the results of models pre-trained on YFCC100M and \jft{} and tested on ImageNet and Places-365~\cite{zhou2017places} with linear evaluation, \ie, features are frozen and a linear classifier is trained. For \jft{} the results are also visualized in Figure~\ref{fig:teaser}. On the ImageNet linear benchmark we see a large drop in performance compared to pre-training on ImageNet (Table %
\ref{tab:linear_baseline}): \showdiffn{-9.0\%}, \showdiffn{-7.3\%} for BYOL-1k and \showdiffn{-9.2\%},  \showdiffn{-7.7\%} for \baseline{}-1k, showing the difficulty of learning representations from uncurated (and more diverse) data.

\input{fig_tex/table_transfer_linear_cls_no_imagenet_v2}

\input{fig_tex/table_finetune_v2}

In these experiments DnC always uses \baseline{}-1k for the clustering, and uses the remaining pre-training epochs for the expert training and distillation. Therefore a good comparison is with \baseline{} trained for longer (from scratch). For 3,000 epochs of training, \baseline{}-3k improves over \baseline{}-1k by \showdiff{+0.6} and \showdiff{+0.8} on YFCC100M and \jft{} respectively, while DnC-3k improves by \showdiff{+2.7} and \showdiff{+3.2}. On Places-365 we see similar relative improvements. We also include BYOL-3k for completeness and again see small improvements with longer training for BYOL. For further longer schedules (4,500-5,000 epochs), we notice similar behavior on both YFCC100M and \jft{}. Besides, we see DnC significantly outperforms concurrent efforts SEER~\cite{goyal2021self} when using ResNet-50. We further test DnC with a larger model (\ie, ResNet-200 with a width multiplier of 2) and observe that DnC outperforms \baseline{} by \showdiff{+3.1}.

\subsection{Transfer Learning}

In this section, we consider both using frozen representations for fine-grained linear classification and fine-tuning for different downstream tasks.

\noindent\textbf{Fine-grained linear classification.}
Following SimCLR~\cite{chen2020simple} and BYOL~\cite{grill2020bootstrap}, we further perform linear classification evaluation on 12 classification datasets (introduced by \cite{kornblith2019better}), to assess whether the learned representation is generic across different image domains (see more details in Section~\ref{sec:finegrain_transfer}). %
As shown in Table~\ref{tbl:transfer_learning}, when pre-training on YFCC100M or \jft{}, DnC significantly and consistently outperforms BYOL and \baseline{}. %

\noindent\textbf{Detection, segmentation, and depth estimation.} In Table~\ref{tbl:transfer_finetune}, we evaluate the representation on three different fine-tuning tasks: (1) for object detection and instance segmentation on COCO~\cite{lin2014microsoft}, we train a standard Mask-RCNN~\cite{he2017mask} using FPN~\cite{lin2017feature} with a 1$\mathbf{x}$ schedule, \ie, 12 epochs; (2) for semantic segmentation on VOC2012, we used FCN~\cite{long2015fully} as in~\cite{he2020momentum}; (3) for depth estimation on NYU-v2 dataset~\cite{silberman2012indoor}, the setup is the same as~\cite{grill2020bootstrap}. In all three tasks, DnC significantly outperforms ImageNet supervised pre-training, \eg, \showdiff{+2.2} in \apbbox{~} and \showdiff{+1.8} in \apmask{~} for detection, \showdiff{+2.5} in mIoU for segmentation, and \showdiff{+5.0} in $<$1.25 metric for depth prediction. DnC also significantly outperforms both self-supervised baselines when transferring to PASCAL and COCO tasks, and performs on-par with MoCLR while outperforming BYOL for depth estimation.

For more implementation details, please refer to Section~\ref{sec:more_transfer} in the Appendix; Complete results for transfer learning are included in Section~\ref{sec:complete_transfer}.

%% file: fig_tex/table_uncurated_v4.tex
\begin{table}[t]
\caption{Comparison of self-supervised learning methods pre-trained on uncurated datasets YFCC100M and \jft{}. For evaluation a linear classifier is trained on ImageNet and Places-365. Computation is measured as ImageNet-equivalent epochs.}
\label{tab:uncurated_results}
\setlength{\tabcolsep}{4pt}
\setlength{\extrarowheight}{-0.5pt}
\begin{center}
\begin{small}
\begin{tabular}{lcccc}
\toprule
Method & Arch & pre-training & ImageNet & Places 365 \\
 & & \# epochs & Top-1 Acc & Top-1 Acc \\
\midrule 
\multicolumn{5}{l}{\emph{Concurrent work trained on IG 1B images:}} \\ %
SEER~\cite{goyal2021self} & R-50  & $\approx$1,000 & 61.6 & - \\
 & R-101 & $\approx$1,000 & 65.8 & - \\
\midrule 
\multicolumn{5}{l}{\emph{Pre-training on YFCC100M:}} \\ %

\baseline{} & \multirow{2}{*}{R-50} & 1,000 & 65.1 & 53.2 \\
BYOL & & 1,000 & 65.3 & 52.9 \\
\specialrule{0em}{2pt}{2pt}
\baseline{} & \multirow{3}{*}{R-50} & 3,000 & 65.7 & 53.2 \\
BYOL &  & 3,000 & 66.6 & 52.9 \\
 DnC &  & 3,000 & \cellcolor{DnCBG}\textbf{67.8} & \cellcolor{DnCBG}\textbf{54.1} \\
\specialrule{0em}{2pt}{2pt}
\baseline{} & \multirow{3}{*}{R-50}& 5,000 & 66.1 & 53.5 \\
BYOL &  & 5,000 & 67.0 & 53.2 \\
 DnC &  & 4,500 & \cellcolor{DnCBG}\textbf{68.5} & \cellcolor{DnCBG}\textbf{54.4} \\
\midrule

\multicolumn{5}{l}{\emph{Pre-training on \jft{}:}} \\

\baseline{} & \multirow{3}{*}{R-50} & 1,000 & 66.6 & 52.1 \\
BYOL & & 1,000 & 67.0 & 51.9 \\
 DnC & & 1,000 & \cellcolor{DnCBG}\textbf{67.9} & \cellcolor{DnCBG}\textbf{52.5} \\
\specialrule{0em}{2pt}{2pt}
\baseline{} & \multirow{3}{*}{R-50} & 3,000  & 67.4 & 52.5 \\
BYOL & & 3,000 & 67.6 & 52.4 \\
 DnC & & 3,000 & \cellcolor{DnCBG}\textbf{69.8} & \cellcolor{DnCBG}\textbf{53.3} \\
\specialrule{0em}{2pt}{2pt}
\baseline{} & \multirow{3}{*}{R-50}& 5,000 & 67.6 & 52.4 \\
BYOL & & 5,000 & 67.9 & 52.4 \\
 DnC & & 4,500 & \cellcolor{DnCBG}\textbf{70.7} & \cellcolor{DnCBG}\textbf{53.5} \\
\specialrule{0em}{2pt}{2pt}
\multicolumn{4}{l}{\emph{With larger ResNet:}} \\
\baseline{} & \multirow{2}{*}{R-200x2}& 3,000 & 74.2 & 54.6 \\
 DnC & & 3,000 & \cellcolor{DnCBG}\textbf{77.3} & \cellcolor{DnCBG}\textbf{56.2} \\

\bottomrule
\end{tabular}
\end{small}
\end{center}
\end{table}

%% file: fig_tex/table_transfer_linear_cls_no_imagenet_v2.tex
\begin{table*}[th]
  \setlength{\tabcolsep}{0pt}
  \setlength{\extrarowheight}{5pt}
  \renewcommand{\arraystretch}{0.75}
  \centering
  \small
  \caption{Transfer learning experiments. We evaluate models pre-trained on ImageNet, YFCC100M and \jft{} with a linear classifier on 12 downstream classification tasks: Food-101~\cite{bossard2014food}, CIFAR-10/100~\cite{krizhevsky2009learning}, Birdsnap~\cite{berg2014birdsnap}, SUN397~\cite{xiao2010sun}, Stanford Cars~\cite{krause2013collecting}, FGVC Aircraft \cite{maji2013fine}, PASCAL VOC 2007 \cite{everingham2010pascal}, Describable Textures (DTD) \cite{cimpoi2014describing}, Oxford-IIIT Pets \cite{parkhi2012cats}, Caltech-101 \cite{fei2004learning} and Oxford 102 Flowers \cite{nilsback2008automated}.}
  \label{tbl:transfer_learning}
  \begin{tabularx}{\linewidth}{p{0.6cm}p{0.2cm}p{1.8cm}p{0.15cm}Cp{0.15cm}Cp{0.15cm}Cp{0.15cm}Cp{0.15cm}Cp{0.15cm}Cp{0.15cm}Cp{0.15cm}Cp{0.15cm}Cp{0.15cm}Cp{0.15cm}Cp{0.15cm}Cp{0.15cm} | Cp{0.01cm}C}
    \toprule
    \\
    \\
    && && \rot{Food-101} && \rot{CIFAR10} && \rot{CIFAR100} && \rot{Birdsnap} && \rot{SUN397} &&  \rot{Cars} &&  \rot{Aircraft}  && \rot{VOC2007} && \rot{DTD}  &&  \rot{Pets} &&  \rot{Caltech-101} &&  \rot{Flowers} && \rot{\textbf{Average}} \\

    \midrule
    \multirow{3}{=}{\centering \rotatebox{90}{YFCC}}
    && BYOL-5k          && 69.1	&& 85.8	&& 66.8	&& \textbf{35.5}	&& 64.1	&& 50.1	&& \textbf{51.9}	&& 82.5	&& 74.5	&& 74.0	&& \textbf{87.6}	&& 95.8	&& 69.8 \\
    && \baseline{}-5k   && 68.4	&& 87.6	&& 69.7	&& 30.5	&& 63.9	&& 41.0	&& 46.7	&& 82.4	&& 76.2	&& 68.5	&& 86.0	&& 93.0	&& 67.8 \\
    && DnC-4.5k           && \textbf{72.1} && \textbf{88.0} && \textbf{71.1} && \textbf{35.5} && \textbf{67.2} && \textbf{52.6} && 49.2 && \textbf{83.7} && \textbf{76.5} && \textbf{75.9} && 87.0 && \textbf{97.8} && \cellcolor{DnCBG}\textbf{71.4} \\

    \midrule
    \multirow{3}{=}{\centering \rotatebox{90}{\jft{}}}
    && BYOL-5k          && 73.3 && 89.8 && 72.4 && 38.2 && 61.8 && 64.4 && \textbf{54.4} && 81.3 && 75.5 && 77.0 && 90.1 && 94.3 && 72.7 \\
    && \baseline{}-5k   && 72.8 && 90.7 && 72.5 && 33.8 && 62.2 && 60.6 && 50.9 && 81.9 && 75.3 && 75.8 && 89.5 && 93.8 && 71.7  \\
    && DnC-4.5k      && \textbf{78.7} && \textbf{91.7} && \textbf{74.9} && \textbf{42.1} && \textbf{65.0} && \textbf{75.3} && 54.1 && \textbf{83.1} && \textbf{76.6} && \textbf{86.1} && \textbf{90.2} && \textbf{98.2} && \cellcolor{DnCBG}\textbf{76.3} \\
    \bottomrule
  \end{tabularx}
\end{table*}

%% file: fig_tex/table_finetune_v2.tex
\begin{table*}[th]
  \setlength{\tabcolsep}{5.8pt}
  \setlength{\extrarowheight}{5pt}
  \renewcommand{\arraystretch}{0.75}
  \centering
  \small
  \caption{Fine-tuning pre-trained model for transfer learning experiments, including object detection on COCO dataset, semantic segmentation on Pascal VOC 2012, and depth estimation on NYU v2 dataset. For the evaluation metrics of \emph{rms} and \emph{rel}, lower is better.}
  \label{tbl:transfer_finetune}

  \begin{tabular}{ll|cccccc|c|ccccc}
  \toprule
  & & \multicolumn{3}{c}{COCO detection} & \multicolumn{3}{c|}{COCO instance seg.} & PASCAL seg. & \multicolumn{5}{c}{NYU v2 depth estimation}\\
  
  & & \apbbox{~} & \apbbox{50} & \apbbox{75} & \apmask{~} & \apmask{50} & \apmask{75} & mIoU & $<$1.25 & $<$1.25$^2$ & $<$1.25$^3$ & rms$\mathcolor{red}{\downarrow}$ & rel$\mathcolor{red}{\downarrow}$ \\
  \midrule
  
  \multicolumn{2}{c|}{ImageNet Super.} & 39.5 & 60.1 & 43.3 & 35.4 & 56.9 & 38.1 & 74.4 & 81.1 & 95.3 & 98.8 & 0.573 & 0.127 \\

  \midrule
  
  \multirow{3}{*}{\small \rotatebox{90}{YFCC}}
  & BYOL-5k & 41.1 & 62.0 & 45.1 & 36.6 & 58.6 & 38.9 & 75.5 & 83.5 & 96.4 & 99.0 & 0.558 & 0.130 \\
  & \baseline-5k & 40.8 & 61.7 & 44.8 & 36.6 & 58.5& 39.0 & 75.1 & \textbf{86.7} & \textbf{97.4} & \textbf{99.3} & \textbf{0.503} & \textbf{0.117} \\
  & DnC-4.5k & \cellcolor{DnCBG}\textbf{41.5} & \cellcolor{DnCBG}\textbf{62.5} & \cellcolor{DnCBG}\textbf{45.6} & \cellcolor{DnCBG}\textbf{37.0} & \cellcolor{DnCBG}\textbf{59.3} & \cellcolor{DnCBG}\textbf{39.6} & \cellcolor{DnCBG}\textbf{76.6} & \cellcolor{DnCBG}86.2 & \cellcolor{DnCBG}97.2 & \cellcolor{DnCBG}\textbf{99.3} & \cellcolor{DnCBG}0.512 & \cellcolor{DnCBG}0.121\\
  
  \midrule
  \multirow{3}{*}{\small \rotatebox{90}{\jft{}}}
  & BYOL-5k & 40.6 & 61.2 & 44.3 & 36.2 & 58.1 & 38.8 & 75.8 & 84.4 & 96.5 & 99.0 & 0.544 & 0.129 \\
  & \baseline-5k  & 41.1 & 62.0 & 45.4 & 36.9 & 58.9 & 39.5 & 76.1 & \textbf{86.3} & \textbf{97.2} & 99.3 & 0.513 & 0.120\\
  & DnC-4.5k & \cellcolor{DnCBG}\textbf{41.7} & \cellcolor{DnCBG}\textbf{62.5} & \cellcolor{DnCBG}\textbf{45.9} & \cellcolor{DnCBG}\textbf{37.2 }& \cellcolor{DnCBG}\textbf{59.3} & \cellcolor{DnCBG}\textbf{39.8} & \cellcolor{DnCBG}\textbf{76.9} & 
  \cellcolor{DnCBG}86.1 & \cellcolor{DnCBG}\textbf{97.2} & \cellcolor{DnCBG}\textbf{99.4} & \cellcolor{DnCBG}\textbf{0.509} & \cellcolor{DnCBG}\textbf{0.119}\\
  
  \bottomrule
  \end{tabular}
\end{table*}

%% file: tex/experiments_imagenet.tex
\section{Hypothesis and Analysis}

The Divide and Contrast (DnC) method hinges on two main hypotheses. The first hypothesis is that clustering activations of powerful self-supervised learning models should provide us with
\emph{locally consistent}
clusters of images (\eg having similar class labels). %
The second is that contrasting against similar (but different) object categories allows self-supervised methods to learn more fine-grained, discriminative representations. 

We empirically assess these hypotheses in isolation. Next we compare DnC with current state of the art methods on ImageNet to see how well it performs on standard (curated) datasets, and analyze the design choices of DnC.

\subsection{Clustered Representations are Object Categories}
Our first hypothesis is that clusters of self-supervised representations are semantically meaningful. To this effect we cluster the representations of various self-supervised learning methods trained on ImageNet with k-means. Specifically, we consider representations from three different layers: the \texttt{pool} layer right after the mean pooling, the \texttt{hidden} layer of the projection head, and the final \texttt{projection}.

We start with 1000-way clustering, and assign
every cluster to a single ImageNet class with a simple majority vote to measure the Top-1 Accuracy. We also measure the mutual information between the clustering assignments and the class labels. Table \ref{tab:clustering_results} gives an overview of these results.
In particular, these methods can group images from the same category surprisingly well, with some representations achieving over 50\% Top-1 clustering accuracy. %
We also notice that the \texttt{hidden} layer performs the best for all methods, and thus use this layer for clustering in the DnC method.

\begin{figure*}[t]
  \centering
  \includegraphics[width=0.96\linewidth]{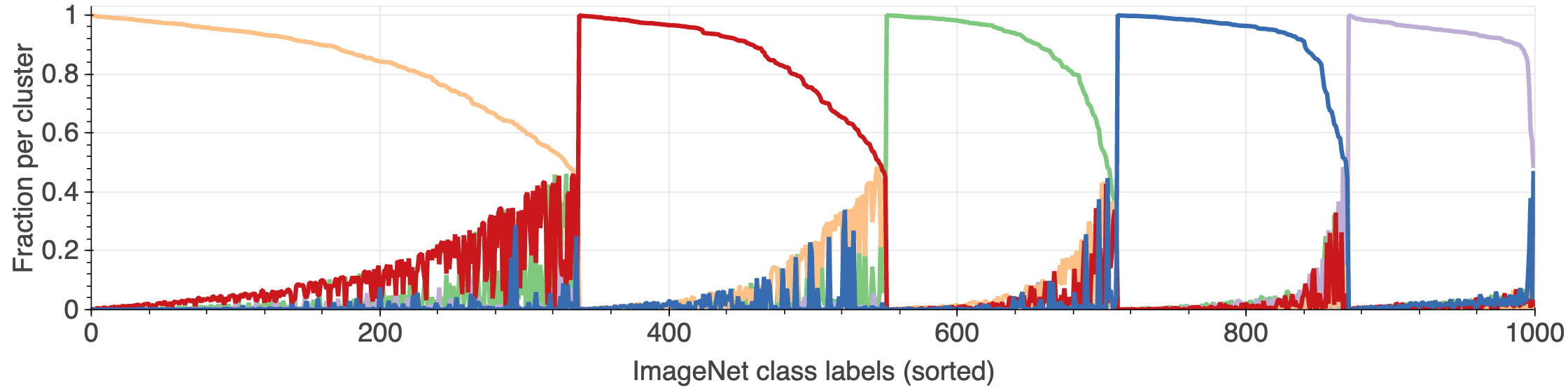}
  \caption{Visualization of the 5-way ImageNet clustering used in DnC. For each ImageNet class we compute the fraction of images that belong to each cluster. For better visualization the x-axis was sorted per cluster. From the figure it is clear that most images in each class belong to a single cluster. The fraction of images in each class that belong to the same cluster is \textbf{87.4\%}.%
  }
  \label{fig:cluster_figure}
\vspace{-5pt}
\end{figure*}

\begin{table}[t]
\caption{Evaluating clustered representations from self-supervised learning methods using Top-1 accuracy and Mutual Information (MI) with the class labels. The representations of each method are clustered using k-means with 1000 centroids. To compute Top-1 accuracy, every cluster is mapped to the most frequent class label for the images in that cluster.}
\label{tab:clustering_results}
\begin{center}
\begin{small}
\vspace{-6pt}
\begin{tabular}{lccccc}
\toprule
Method & Layer & Dimension & Top-1 Acc & MI \\
\midrule 
\multirow{3}{*}{SimCLR~\cite{chen2020simple}} 
& pool & 2048 & 30.1 & 5.64 \\
& hidden & 2048 & 33.3 & 5.83 \\
& output & 128 & 31.8 & 5.68 \\ 
\midrule 
\multirow{3}{*}{BYOL~\cite{grill2020bootstrap}} 
& pool & 2048 & 39.9 & 6.23 \\
& hidden & 4096 & 51.0 & 6.99 \\
& output & 256 & 50.0 & 6.78 \\ 
\midrule 
\multirow{3}{*}{\baseline{} (ours)} 
& pool & 2048 & 40.1 & 6.26 \\
& hidden & 4096 & \textbf{51.6} & \textbf{7.19} \\
& output & 256 & 49.8 & 7.11 \\ 
\bottomrule
\end{tabular}
\end{small}
\end{center}
\vskip -0.1in
\vspace{-8pt}
\end{table}

To give an orthogonal view with a smaller number of clusters, Figure \ref{fig:cluster_figure} plots the 5 clusters used in the DnC model for ImageNet (based on the clustering of a \baseline{} ResNet-50).
Qualitatively, it appears that groups of classes are jointly assigned to the same cluster. Indeed, 
the fraction of images in each class that belong to the same cluster is 87.4\%, lending further evidence that individual clusters are semantically coherent.

\begin{table}[t]
\caption{Linear classification evaluated on the Canine subset of ImageNet (130 classes). The feature extraction models were pre-trained either on ImageNet (Full) or the subset of Canine images without using any labels. All computation is reported in terms of ``full imagenet" epochs. We report the difference in performance relative to training on the full dataset for 1,000 epochs. Even though the canine-only models were trained with 5$\times$ fewer gradient updates, they largely outperform the self-supervised learning models that were trained on the full dataset. We also observe that contrastive methods (SimCLR, \baseline{}) benefit the most.}
\vspace{-4pt}
\label{tab:canine_results}
\begin{center}
\begin{small}
\begin{tabular}{lcccc}
\toprule
Method & Pre-training Dataset & Epochs  & Top-1 Acc \\
\midrule 
SimCLR & Full & \hspace{0.5em} 200  & 67.4  \\
BYOL & Full & \hspace{0.5em} 200  & 70.7  \\
\baseline{} & Full & \hspace{0.5em} 200  & 68.7  \\ 
\midrule 
SimCLR & Full & 1,000 & 69.4 \\
BYOL & Full & 1,000 & 76.5  \\
\baseline{} & Full & 1,000 & 75.3   \\ 
\midrule 
SimCLR & Canine & \hspace{0.25em} 100 & 72.8 \showdiff{(+3.4)}  \\
BYOL & Canine & \hspace{0.25em} 100 & 76.0 \textbf{(-0.5)}  \\
\baseline{} & Canine & \hspace{0.25em} 100  & 76.1 \showdiff{(+0.8)}  \\
\midrule 
SimCLR & Canine & \hspace{0.5em}  200 & 74.1 \showdiff{(+4.7)}  \\
BYOL & Canine & \hspace{0.5em} 200 & 77.3 \showdiff{(+0.8)}  \\
\baseline{} & Canine & \hspace{0.5em} 200 & \textbf{77.5} \showdiff{(+2.2)} \\
\bottomrule
\end{tabular}
\end{small}
\end{center}
\vspace{-10pt}
\end{table}

\subsection{Training on Semantically Similar Data Subsets}
\label{section:canine_experiments}
DnC is based on the second hypothesis that training self-supervised learning methods on a subset of images from similar object categories should improve performance on those object classes. 
Contrastive methods in particular stand to benefit from this procedure, as distinguishing positive samples from negatives from nearby classes might require learning more fine-grained features 
(similarly to hard-negative mining~\cite{schroff2015facenet,kalantidis2020hard}). On the other hand, it might hinder performance by drawing negative samples that are too similar, including more false-negatives \cite{chuang2020debiased}. 

To test this hypothesis in isolation and gain a better intuitive understanding of our method, we train various self-supervised learning models on the subset of ImageNet classes that belong to the canine family (including dogs, wolves and foxes, 130 classes in total) and compare them to models trained on the full dataset. For all models, we train a linear classifier on the canine-only subset, and evaluate on validation images from the canine subset.

From Table~\ref{tab:canine_results} it can be seen that models pre-trained on the canine-only subset perform significantly better than those trained on the entire ImageNet dataset, even though they have significantly less images to learn from and were trained with 5$\times$ less computation.

\subsection{ImageNet Results}
Even though our main goal is to improve self-supervised learning on uncurated datasets, we asked whether it remains competitive on heavily studied datasets such as ImageNet.

\input{fig_tex/table_imagenet_epochs}

From Table~\ref{tab:imagenet_results} we see that training the baseline \baseline{} for 2,000 more epochs does not improve the results by much (+0.2). DnC on the other hand convincingly outperforms the baseline (\showdiff{+1.3}). Interestingly, DnC even slightly outperforms \baseline{} or BYOL when giving a computational budget of 1,000 epochs. Though DnC aims at uncurated datasets, the previous results from Section \ref{section:canine_experiments} on the canine subset have shown that even on ImageNet it might be beneficial to draw negatives from more similar categories.

\input{fig_tex/table_semisup}
\subsubsection{Semi-supervised learning}
We evaluate the performance of DnC when fine-tuning on a subset of ImageNet's train set. Following the semi-supervised protocol~\cite{kornblith2019better,zhai2019s4l,chen2020simple,grill2020bootstrap}, we use the same splits of $1\%$ and $10\%$ ImageNet data as in~\cite{chen2020simple,grill2020bootstrap}. As shown in Table~\ref{tab:semi_results}, DnC consistently outperforms BYOL, SwAV, and MoCLR and Barlow Twins. 

\begin{table}[ht]
\caption{Ablating the experts used in DnC: We notice a big drop in performance if the experts models are trained on the whole dataset (ensemble), or on random subsets.}
\vspace{-5pt}
\label{tab:ablation_expert}
\begin{center}
\begin{small}
\setlength{\tabcolsep}{4pt}
\begin{tabular}{cccc}
\toprule
& Partitioning & Experts trained on  & Top-1 Acc \\
\midrule 
DnC & Clustering & local partition          & 75.8  \\
- local experts & - & full dataset & 74.3  \\
- clustering &Randomly     & local partition  & 73.1  \\
\bottomrule
\end{tabular}
\end{small}
\end{center}

\caption{Evaluation of what models to predict during distillation.}
\vspace{-4pt}
\setlength{\tabcolsep}{4pt}
\label{tab:complementary}
\begin{center}
\begin{small}
\begin{tabular}{cccc}
\toprule
base model & local experts & use center-crop  & Top-1 Acc \\
\midrule 
\checkmark &            &    &  74.5 \\
          & \checkmark &    &  75.2 \\
\checkmark & \checkmark &    &  \textbf{75.8} \\
\hline
\checkmark & \checkmark &  \checkmark  &  75.6 \\
\bottomrule
\end{tabular}
\end{small}
\end{center}
\vskip -0.1in
\end{table}

\subsubsection{Ablations}
We provide further experiments for isolating the factors that make DnC work, shown in Table~\ref{tab:ablation_expert}. If we train the expert models on the full dataset instead of subsets (similar to an ensemble), but with the same computational budget, the resulting model achieves the same performance as the base model (no improvement). Alternatively, splitting the dataset into random subsets hurts the final performance, showing the importance of the clustering used.

In Table~\ref{tab:complementary} we study the distillation process, showing it is important to regress to both the base model and experts. And as discussed in Section~\ref{section:distill}, using center-crops instead of augmented views does not hurt the performance by much.

%% file: fig_tex/table_imagenet_epochs.tex
\begin{table}[h]
\caption{Comparison with BYOL and MoCLR on ImageNet linear evaluation benchmark with training budgets of 1000 and 3000 epochs. Top-1 accuracy is reported using ResNet-50.
}
\vspace{-4pt}
\label{tab:imagenet_results}
\begin{center}
\begin{small}
\begin{tabular}{lccc}
\toprule

Method & 1000 epochs & 3000 epochs & $\Delta$ \\
\midrule
BYOL        & 74.3 & 73.9 & -0.4 \\
MoCLR       & 74.3 & 74.5 & 0.2  \\
DnC (ours)  & \cellcolor{DnCBG}\textbf{74.5} & \cellcolor{DnCBG}\textbf{75.8} & \cellcolor{DnCBG}\textbf{1.3}  \\
\bottomrule

\end{tabular}
\end{small}
\end{center}
\vspace{-15pt}
\end{table}

%% file: fig_tex/table_semisup.tex
\begin{table}[t]
\caption{Semi-supervised results with a fraction of ImageNet labels following the protocol of~\cite{chen2020simple,grill2020bootstrap}. The encoder is ResNet-50.}
\vspace{-5pt}
\label{tab:semi_results}
\setlength{\tabcolsep}{4.5pt}
\begin{center}
\begin{small}
\begin{tabular}{lcccccc}
\toprule
 & \multicolumn{3}{c}{Top-1} & \multicolumn{3}{c}{Top-5}\\
method & \multicolumn{3}{c}{Label fraction} & \multicolumn{3}{c}{Label fraction}\\
 & $1\%$ & $10\%$ & $100\%$ & $1\%$ & $10\%$ & $100\%$ \\
\midrule
SimCLR~\cite{chen2020simple}     & 48.3 & 65.6 & 76.0    & 75.5 & 87.8 & 93.1 \\
BYOL~\cite{grill2020bootstrap}       & 53.2 & 68.8 & 77.7    & 78.4 & 89.0 & 93.9 \\
SwAV~\cite{caron2020unsupervised}    & 53.9 & 70.2 & -    & 78.5 & 89.9 & - \\
\baseline{}& 53.0 & 68.8 & 77.4    & 79.1 & 89.6 & 94.0 \\
Barlow Tw.~\cite{zbontar2021barlow} & 55.0 & 69.7 & - & 79.2 & 89.3 & - \\ 
DnC        & \cellcolor{DnCBG}\textbf{59.9} & \cellcolor{DnCBG}\textbf{71.1} & \cellcolor{DnCBG}\textbf{78.2} & \cellcolor{DnCBG}\textbf{83.0} & \cellcolor{DnCBG}\textbf{90.4} & \cellcolor{DnCBG}\textbf{94.2} \\
\bottomrule
\end{tabular}
\end{small}
\end{center}
\vspace{-15pt}
\end{table}

%% file: tex/conclusion.tex
\section{Conclusion}
In this paper we have studied how state of the art self-supervised learning methods perform when they are pretrained on uncurated data -- datasets that did not require human annotations or labels to create -- as a step towards fully self-supervised learning. We have observed that current methods suffer from a large drop in performance of up to \showdiffn{-9\%} when pre-trained on these uncurated datasets. To alleviate this issue, we have proposed Divide and Contrast (DnC) that requires a few simple changes to existing self-supervised learning methods, and which largely outperforms state of the art SSL methods on uncurated datasets, as well as achieving similar or better performance on ImageNet. %
We hope this work draws more attention to %
uncurated datasets as a benchmark for self-supervised learning.

\vspace{5pt}
\noindent \textbf{Acknowledgements.} We are grateful to Florent Altché, Bilal Piot, Jean-Bastien Grill, Elena Buchatskaya, and Florian Strub for significant help with reproducing BYOL results; Jeffrey De Fauw for providing the initial code base for SimCLR; Carl Doersch, Lucas Beyer, Phillip Isola, and Oriol Vinyals for valuable feedback on the manuscript.

%% file: tex/sup.tex
\section{Image Augmentations}
\label{app:transf}

For a fair comparison, we used exactly the same image augmentations as BYOL~\cite{grill2020bootstrap} (which are a subset of the ones presented in SimCLR~\cite{chen2020simple}):
\begin{itemize}
    \item random resized cropping: a random patch is cropped, whose area is uniformly sampled between $0.08\times$ and $1\times$ that of the raw image, and aspect ratio is logarithmically sampled between $3/4$ and $4/3$. We resize the patch to $224 \times 224$ pixels using bicubic interpolation;
    \item random horizontal flip;
    \item color jittering: the brightness, contrast, saturation and hue of the image are shifted by a uniformly distributed offset applied on all the pixels of the same image;
    \item color dropping: randomly convert images to grayscale, computed as $0.2989 R + 0.5870 G + 0.1140 B$;
    
    \item Gaussian blurring: a Gaussian kernel of size $23 \times 23$ is used, whose standard deviation is uniformly sampled from $[0.1, 2.0]$;
    \item solarization: an optional color transformation $x \mapsto x \cdot \bOne_{\{x < 0.5\}} + (1 - x) \cdot \bOne_{\{x \ge 0.5\}}$ for pixels with values in $[0, 1]$.
\end{itemize}

Augmentations from the sets $\mathcal{T}$ and~$\mathcal{T}'$ are compositions of the above image augmentations, each applied with a predetermined probability. The parameters for $\mathcal{T}$ and~$\mathcal{T}'$ are listed in Table~\ref{tab:transformation_distributions}.

In the evaluation or representation clustering stage, we follow the standard center-crop strategy: resize images to $256$ pixels along the shorter side, and crop out the central $224 \times 224$ window.

\begin{table}[ht]
    \small
    \centering
    \begin{tabular}{l l l} \toprule
        Parameter & $\mathcal{T}$ & $\mathcal{T}'$ \\ \midrule
        Random crop probability & $1.0$ & $1.0$ \\
        Flip probability & $0.5$ & $0.5$ \\
        Color jittering probability & $0.8$ & $0.8$ \\
        Brightness adjustment max intensity & $0.4$ & $0.4$ \\
        Contrast adjustment max intensity & $0.4$ & $0.4$ \\
        Saturation adjustment max intensity & $0.2$ & $0.2$ \\
        Hue adjustment max intensity & $0.1$ & $0.1$ \\
        Color dropping probability & $0.2$ & $0.2$ \\
        Gaussian blurring probability & $1.0$ & $0.1$ \\
        Solarization probability & $0.0$ & $0.2$ \\ \bottomrule
    \end{tabular}
     \vspace{0.5em}
    \caption{Parameters used to generate image augmentations during training, which are exactly the same as those in~\cite{grill2020bootstrap}.}
    \label{tab:transformation_distributions}
\end{table}

\section{Pre-trainning Datasets}

\noindent\textbf{ImageNet.} We split out $10009$ images from the train set as our local validation set, and use the remaining $1271158$ images for both unsupervised pre-training and linear classifier training. After selecting hyper-parameters based on the performance of the local validation set, we report accuracy on the official validation set consisting of $50000$ images.

\noindent\textbf{\jft{}.} The \jft{} dataset contains $301.7$ millions of images in total.

\noindent\textbf{YFCC100M.} YFCC-100M is a widely used uncurated dataset that includes $\sim$95 millions of images, which are all used in our pre-training. 

\section{Clustering Representations}
We apply the vanilla k-means algorithm on the representations extracted from the \texttt{hidden} layer of the projection network, with cosine similarity as a distance metric. When pre-training on ImageNet, we use all training images for clustering; when pre-training on \jft{} and YFCC, we randomly sample $1.5$ million images for clustering and extracting the centroids, and then use these centroids to assign clustering labels to all images. 

\section{Run Time Analysis of DnC}\label{sec:run_time}

While DnC has three stages of training, its computational complexity or running time is similar as other state-of-the-art approaches trained for the same number of epochs. As a illustration, we compare the training FLOPs of DnC with other methods such as BYOL, \baseline{}, and SwAV. As discussed in the main paper, the base or expert training stage of DnC has exactly the same FLOPs as the chosen base approach, \ie, \baseline{} here. The only different lies in the third stage, where DnC requires one additional forward pass. Therefore, for DnC we compute a weighted average of FLOPs across three stages (we use the normalized number of training epochs as weights). Table~\ref{tab:flops} summarizes the comparison with other approaches: DnC is comparable with BYOL and \baseline{}, while SwAV costs more flops because it uses eight views per image per step.

\begin{table}[t]
\caption{Training FLOPs of different methods per image when using ResNet-50.}
\label{tab:flops}
\setlength{\tabcolsep}{4.5pt}
\begin{center}
\begin{small}
\begin{tabular}{c|cccc}
\toprule
               & SwAV  & BYOL  & MoCLR & DnC \\
\midrule
training FLOPS & 38.4B & 24.7B & 24.7B & 25.4B \\
\bottomrule
\end{tabular}
\end{small}
\end{center}
\end{table}

Besides, we also run BYOL, MoCLR and DnC on ImageNet for 3000 epochs to compare the running time. Table~\ref{tab:time} reports the comparison when using 512 TPU v3 cores. DnC only introduces $<$5\% extra training time, compared with BYOL and MoCLR. Besides, the time cost for clustering the representations is small, \eg, it takes about $20$-$30$ minutes to extract representations on the training set and cluster them into groups, even only with $8$ V100 GPUs on a single node.

\begin{table}[h]
\caption{Training time for 3000 epochs on ImageNet when using 512 TPU v3 cores.}
\label{tab:time}
\setlength{\tabcolsep}{4.5pt}
\begin{center}
\begin{small}
\begin{tabular}{c|ccc}
\toprule
                     & BYOL  & MoCLR & DnC \\
\midrule
training time (hours) & $\sim$24 & $\sim$24 & $\sim$25 \\
\bottomrule
\end{tabular}
\end{small}
\end{center}
\end{table}

\section{Optimization}

\noindent\textbf{Unsupervised pre-training.} All the hyper-parameters for optimization directly follow BYOL, except for base learning rate (which we discuss in the next paragraph). Specifically, we use LARS optimizer~\cite{LARS} with a cosine decay learning rate schedule~\cite{SGDR} and a warm-up period of $10$ epochs for all unsupervised pre-training. 
In addition, we use a global weight decay parameter of $1.5\cdot10^{-6}$ while excluding the biases and batch normalization parameters from both LARS adaptation and weight decay. For the momentum encoder, its parameters $\theta_\text{EMA}$ are updated by $\theta_\text{EMA}=\tau\theta_\text{EMA} + (1-\tau)\theta$, where $\theta$ are the parameters of the online encoder. The EMA parameter $\tau$ starts from $\tau_\text{base} = 0.996$ and is increased to one during training. Following BYOL, we set 
\vspace{-3pt}
\begin{equation}\label{eq:tau}
\tau \delequal 1 - (1-\tau_\text{base})\cdot \pa{\cos\pa{\pi k / K} + 1} / 2
\end{equation}
\vspace{-3pt}
with $k$ the current training step and $K$ the maximum number of training steps. 

Specifically for the base learning rate, we used $0.2$ for BYOL as in the original paper (we sweep over $\{0.2,0.3,0.4\}$ for 1000 epochs pre-training on ImageNet to confirm $0.2$ is the best). For \baseline{}, we found $0.3$ is slightly better than $0.2$, and therefore we kept using $0.3$ for \baseline{} and all stages of DnC (The only exception is that for DnC with 1000 epoch schedule, we increase the base learning rate to 0.5 to compensate for short training of models at each stage). The final learning rate is scaled linearly~\cite{Goyal2017} with the batch size ($\text{LearningRate} = \text{BaseLearningRate} \times \text{BatchSize} / 256$).

\noindent\textbf{Linear evaluation on ImageNet/Places-365.} On top of the global pooling layer of the frozen pre-trained encoder, we train a supervised 1000- or 365-way linear classifier, as in~\cite{zhang2016colorful,tian2019contrastive,he2020momentum,chen2020simple}. We optimize the cross-entropy loss using SGD with Nesterov momentum over $80$ epochs, using a batch size of $1024$ and a cosine learning rate decay schedule. We sweep the base learning rate (of batch size 256) over $\{0.4, 0.3, 0.2, 0.1, 0.05\}$ for models pre-trained on ImageNet, and $\{1.0, 0.6, 0.4, 0.2, 0.1\}$ for models pre-trained on \jft{} and YFCC. We chose the best learning rate on a local validation set split out from the ImageNet train set, and report the accuracy on the official ImageNet validation set.

\section{Transfer to Other Datasets}\label{sec:more_transfer}

\subsection{Implementation: fine-grained linear classificaton}\label{sec:finegrain_transfer}

We perform transfer via linear classification and fine-tuning on the same set of datasets as in~\cite{chen2020simple,grill2020bootstrap}, namely Food-101~\cite{bossard2014food}, CIFAR-10/100~\cite{krizhevsky2009learning}, Birdsnap~\cite{berg2014birdsnap}, SUN397~\cite{xiao2010sun}, Stanford Cars~\cite{krause2013collecting}, FGVC Aircraft \cite{maji2013fine}, PASCAL VOC 2007 classification task~\cite{everingham2010pascal}, Describable Textures (DTD) \cite{cimpoi2014describing}, Oxford-IIIT Pets \cite{parkhi2012cats}, Caltech-101 \cite{fei2004learning} and Oxford 102 Flowers \cite{nilsback2008automated}. As in~\cite{chen2020simple,grill2020bootstrap}, we used the validation sets specified by the dataset creators to select hyperparameters for FGVC Aircraft, PASCAL VOC
2007, DTD, and Oxford 102 Flowers. On other datasets, we use the validation examples as test set, and hold out a subset of the training examples as validation set while performing hyperparameter tuning.

We follow the linear evaluation protocol of~\cite{kolesnikov2019revisiting,kornblith2019better,chen2020simple,grill2020bootstrap}. 
We train a regularized multinomial logistic regression classifier on top of the frozen representation without data augmentation. Images are resized to $224$ pixels along the shorter side and cropped by the center $224 \times 224$ pixels.
We minimize the cross-entropy objective using L-BFGS with $\ell_2$-regularization, where we select the regularization parameters from a range of $45$ logarithmically-spaced values between $10^{-6}$ and $10^5$. 

\subsection{Implementation: Pascal VOC segmentation}

Following BYOL, we use the same fully-convolutional network (FCN)-based~\cite{long2015fully} architecture as~\cite{he2020momentum}. The backbone consists of the convolutional layers in ResNet-$50$. The $3\times 3$ convolutions in the conv$5$ blocks use dilation $2$ and stride $1$. This is followed by two extra $3\times 3$ convolutions with $256$ channels, each followed by batch normalization and ReLU activations, and a $1\times 1$ convolution for per-pixel classification. The dilation is set to $6$ in the two extra $3\times 3$ convolutions. The total stride is $16$ (FCN-$16$s \cite{long2015fully}).

Similar as BYOL, we train on the \texttt{train2012\xspace} set and report results on \texttt{val2012\xspace}. Hyperparameters are selected on a $2119$ images held-out validation set. Training is done with random scaling (by a ratio in $\left[0.5, 2.0\right]$), cropping, and horizontal flipping. The crop size is $513$. Inference is performed on the $\left[513, 513\right]$ central crop. For training we use a batch size of $16$ and weight decay of $0.0001$. We select the base learning rate by sweeping across $5$ logarithmically spaced values between $10^{-3}$ and $10^{-1}$. The learning rate is multiplied by $0.1$ at the $70$-th and $90$-th percentile of training. We train for $30000$ iterations, and average the results on 5 seeds.

\subsection{Implementation: COCO detection}

We use the standard Mask R-CNN~\cite{he2017mask} with the FPN~\cite{lin2017feature} backbone, with cross-replica BN tuned, similar as that in MoCo~\cite{he2020momentum}. We fine-tune all layers end-to-end. We finetune on the \texttt{train2017} set ($\sim$118k images) and evaluate on \texttt{val2017}. We use the standard ``$1$x schedule''. 

We directly use the public Cloud TPU implementation without modification\footnote{\small{\url{https://github.com/tensorflow/tpu/tree/master/models/official/detection}}}. Specifically, we use a batch size of 64 images split over 16 workers. We linearly warmup the learning rate to 0.3 for the first 500 iterations, and drop it twice by a factor of 2, after $\frac{2}{3}$ and $\frac{8}{9}$ of the total training steps.

\subsection{Implementation: NYU v2 depth estimation}

Similar as BYOL, we follow the same protocol as in~\cite{laina2016depth}. With a standard ResNet-$50$ backbone, we feed the conv$5$ features into $4$ fast up-projection blocks with respective filter sizes $512$, $256$, $128$, and $64$. We use a reverse Huber loss function for training.

The original NYU Depth v$2$ frames of size $\left[640, 480\right]$ are down-sampled by a factor $0.5$ and center-cropped to $\left[304, 228\right]$ pixels. Input images are randomly horizontally flipped and the same set of color transformations as in~\cite{grill2020bootstrap} are applied. We train for $7500$ steps with batch size $256$, weight decay $0.0005,$ and learning rate $0.16$ (scaled linearly from the setup of \cite{laina2016depth} to account for the larger batch size).

\section{DnC with other Self-supervised Methods}
While the main paper demonstrate the effectiveness of DnC with \baseline{}, we found DnC can potentially improves other state-of-the-art self-supervised approaches as well. In Table~\ref{tab:combination}, we demonstrate that DnC can improve SimCLR significantly and also benefit BYOL, when both pre-training and evaluating on ImageNet.

\begin{table}[h]
\caption{Applying DnC with other self-supervised methods, when pre-training and evaluating on ImageNet.}
\label{tab:combination}
\begin{center}
\begin{small}
\begin{tabular}{c|c|ccc}
\toprule
Method & w/ DnC & Epochs & Accuracy (\%) & $\Delta$ \\
\midrule
SimCLR &            & 1000 & 69.4 & \\
       &            & 5000 & 70.2 & ~\showdiff{+0.8}\\
       & \checkmark & 3000 & 73.0 & ~\showdiff{+3.6}\\
\midrule
BYOL   &            & 1000 & 74.3 & \\
       &            & 3000 & 73.9 & ~\textbf{-0.5}\\
       & \checkmark & 3000 & 75.1 & ~\showdiff{+0.8} \\
\bottomrule
\end{tabular}
\end{small}
\end{center}
\end{table}

Besides, we notice that DnC with BYOL gets a larger improvement when pre-training on the uncurated dataset YFCC. As shwon in Table~\ref{tab:combination_byol}, naively extending BYOL from 1000 to 5000 epochs only increases the performance by $1.7\%$, while DnC-4500 leverages the computation more efficiently and improves the accuracy by $3.4\%$.

\begin{table}[h]
\caption{Applying DnC with BYOL on YFCC pre-training. Evaluation is conducted on ImageNet linear evaluation.}
\label{tab:combination_byol}
\begin{center}
\begin{small}
\begin{tabular}{c|c|ccc}
\toprule
Method & w/ DnC & Epochs & Accuracy (\%) & $\Delta$ \\
\midrule
BYOL   &            & 1000 & 65.3 & \\
       &            & 3000 & 66.6 & ~\showdiff{+1.3} \\
       &            & 5000 & 67.0 & ~\showdiff{+1.7} \\
       & \checkmark & 3000 & 67.9 & ~\showdiff{+2.6} \\
       & \checkmark & 4500 & 68.7 & ~\showdiff{+3.4} \\
\bottomrule
\end{tabular}
\end{small}
\end{center}
\end{table}

\section{Comparing with SoTA on ImageNet}

\input{fig_tex/table_imagenet_complete}

Though ImageNet linear evaluation benchmark (both pre-training and evaluating on ImageNet) is not the main focus of this paper, we still provides a comparison between DnC and recent SoTA methods, as shown in Table~\ref{tab:imagenet_results_complete}.

\section{Additional Ablations and Results}

\subsection{Length of the distillation stage}
While the distillation stage introduces additional FLOPs compared to the base or expert training stage, this stage can be short. In this section, we conduct ablation on the number of epochs for distillation stage. We train DnC on ImageNet following the the DnC-3k schedule (\ie, 1000 epochs for base training and 1500 epochs for experts). We vary the number of epochs used for distillation and report the linear evaluation accuracy in Table~\ref{tab:distill}. Short distillation schedule such as 60 epochs can yield $74.0\%$, as long as a larger learning rate is utilized to compensate for the smaller number of gradient steps.

\begin{table}[h]
\caption{Abalation on length of the distillation stage.}
\label{tab:distill}
\begin{center}
\begin{small}
\begin{tabular}{c|ccccc}
\toprule
Epoch & 60  & 100 & 200 & 300 & 500 \\
\midrule
Learning rate & 0.45 & 0.45 & 0.35 & 0.35 & 0.3 \\
Accuracy ($\%$) & 74.0 & 74.9 & 75.1 & 75.4 & 75.8 \\
\bottomrule
\end{tabular}
\end{small}
\end{center}
\end{table}

\subsection{$\ell_2$-normalization on regressor output}

We study whether it's better to normalize the output of the regressor by $\ell_2$-normalization in the distillation stage. We conducted this ablation on ImageNet, and found normalizing the output of the regressor actually hurts the performance a bit, as shown in Table~\ref{tab:normalize}.

\begin{table}[h]
\caption{Abalation on whether $\ell_2$-normalize the output of regressors $r_b$ and $r_k (k=1,2,...K)$.}
\label{tab:normalize}
\begin{center}
\begin{small}
\begin{tabular}{c|c}
\toprule
$\ell_2$-normalization & Accuracy \\
\midrule
Yes & 75.4 \\
No  & 75.8 \\
\bottomrule
\end{tabular}
\end{small}
\end{center}
\end{table}

\subsection{Semi-supervised learning with projection layer}
As found in SimCLR v2~\cite{chen2020big}, fine-tuning from the hidden layer of projection head gives better semi-supervised accuracy. In this seciton, we also report the semi-supervised accuracy fine-tuned from the hidden layer of the projection head in Table~\ref{tab:semi_results_hidden}. Models are all pre-trained using ImageNet data.

\begin{table}[h]
\caption{Semi-supervised results by fine-tuning from the first layer of the projection head, following \cite{chen2020simple}. The encoder is ResNet-50.}
\label{tab:semi_results_hidden}
\begin{center}
\begin{small}
\begin{tabular}{lcccc}
\toprule
 & \multicolumn{2}{c}{Top-1} & \multicolumn{2}{c}{Top-5}\\
method & \multicolumn{2}{c}{Label fraction} & \multicolumn{2}{c}{Label fraction}\\
 & $1\%$ & $10\%$ & $1\%$ & $10\%$ \\
\midrule
SimCLR v2     & 57.9 & 68.4  & 82.5 & 89.2\\
BYOL          & 61.9 & 71.9  & 83.3 & 90.7 \\
\baseline{}   & 61.0 & 71.6  & 84.2 & 90.9 \\
DnC           & \cellcolor{DnCBG}\textbf{65.6} & \cellcolor{DnCBG}\textbf{73.2}    
              & \cellcolor{DnCBG}\textbf{86.4} & \cellcolor{DnCBG}\textbf{91.4} \\
\bottomrule
\end{tabular}
\end{small}
\end{center}
\end{table}

\input{fig_tex/table_transfer_linear_complete}

\input{fig_tex/table_finetune_complete}

\subsection{Complete results of transfer learning}\label{sec:complete_transfer}

In Table~\ref{tbl:transfer_learning_linear_complete}, we summarize the transfer learning results on fine-grained linear classification tasks, with different computational budgets in pre-training stage for each method.

In Table~\ref{tbl:transfer_finetune_complete}, we provide the complete results of transfer learning on COCO detection, Pascal VOC semantic segmentation, and NYU depth estimation.

%% file: fig_tex/table_imagenet_complete.tex
\begin{table}[t]
\caption{Linear evaluation benchmark on ImageNet (self-supervised pre-training is also conducted on ImageNet). * indicates using eight views for pre-training.
}
\label{tab:imagenet_results_complete}
\begin{center}
\begin{small}
\begin{tabular}{lccc}
\toprule
Method & Epochs & Top-1 Acc & Top-5 Acc \\
\midrule 
\multicolumn{4}{l}{\textit{Clustering methods with ResNet-50:}}\\
SeLa \cite{Asano2020}  & \hspace{1.5ex}400  & 61.5 & 84.0\\
DeepClusterV2* \cite{caron2020unsupervised} & \hspace{1.5ex}800  & 75.2 & - \\
SwAV* \cite{caron2020unsupervised} & \hspace{1.5ex}800  & 75.3 & - \\
\midrule 
\multicolumn{4}{l}{\textit{Contrastive learning with designed architecture:}}\\
AMDIM \cite{bachman2019learning} & \hspace{1.5ex}150 & 68.1 & - \\
CMC \cite{tian2019contrastive} & \hspace{1.5ex}240  & 70.6  & 89.7\\
\multicolumn{4}{l}{\textit{Contrastive learning with ResNet-50:}}\\
NPID \cite{wu2018unsupervised} & \hspace{1.5ex}200  & 56.5  & -    \\
Local Agg. \cite{zhuang2019local} & \hspace{1.5ex}200  & 58.8  & -    \\
CPC v2~\cite{henaff2019data}      & -    & 63.8  & 85.3  \\
MoCo \cite{he2020momentum} & \hspace{1.5ex}200 & 60.6 & - \\
PIRL \cite{misra2019self}         & \hspace{1.5ex}800 & 67.4 & - \\
PCL \cite{li2020prototypical}         & \hspace{1.5ex}200 & 67.6 & - \\
SimCLR \cite{chen2020simple} & 1,000 & 69.3 & 89.0 \\
PIC \cite{cao2020parametric} & 1,600 & 70.8 & 90.0 \\
MoCo v2 \cite{chen2020improved} & \hspace{1.5ex}800 & 71.1 & - \\
SimCLR v2 \cite{chen2020big} & 1,000 & 71.7 & 90.4\\
InfoMin Aug. \cite{tian2020makes} & \hspace{1.5ex}800 & 73.0 & 91.1 \\
BYOL \cite{grill2020bootstrap} & 1,000 & 74.3 & 91.6\\
BYOL \cite{grill2020bootstrap} & 3,000 & 73.9 & 92.2 \\
\baseline{}~(ours)  & 1,000 & 74.3 & 92.2 \\
\baseline{}~(ours)  & 3,000 & 74.5 & 92.3 \\
DnC~(ours)          & 1,000 & \cellcolor{DnCBG}74.5 & \cellcolor{DnCBG}92.2 \\
DnC~(ours) & 3,000 & \cellcolor{DnCBG}\textbf{75.8} & \cellcolor{DnCBG}\textbf{92.8} \\
\bottomrule
\end{tabular}
\end{small}
\end{center}
\end{table}

%% file: fig_tex/table_transfer_linear_complete.tex
\begin{table*}[th]
  \setlength{\tabcolsep}{0pt}
  \setlength{\extrarowheight}{5pt}
  \renewcommand{\arraystretch}{0.75}
  \centering
  \small
  \caption{Transfer learning experiments. We evaluate models pre-trained on ImageNet, YFCC100M and \jft{} with a linear classifier on 12 downstream classification tasks: Food-101~\cite{bossard2014food}, CIFAR-10/100~\cite{krizhevsky2009learning}, Birdsnap~\cite{berg2014birdsnap}, SUN397~\cite{xiao2010sun}, Stanford Cars~\cite{krause2013collecting}, FGVC Aircraft \cite{maji2013fine}, PASCAL VOC 2007 \cite{everingham2010pascal}, Describable Textures (DTD) \cite{cimpoi2014describing}, Oxford-IIIT Pets \cite{parkhi2012cats}, Caltech-101 \cite{fei2004learning} and Oxford 102 Flowers \cite{nilsback2008automated}.}
  \label{tbl:transfer_learning_linear_complete}
  \begin{tabularx}{\linewidth}{p{0.6cm}p{0.2cm}p{1.8cm}p{0.15cm}Cp{0.15cm}Cp{0.15cm}Cp{0.15cm}Cp{0.15cm}Cp{0.15cm}Cp{0.15cm}Cp{0.15cm}Cp{0.15cm}Cp{0.15cm}Cp{0.15cm}Cp{0.15cm}Cp{0.15cm} | Cp{0.01cm}C}
    \toprule
    \\
    \\
    && && \rot{Food-101} && \rot{CIFAR10} && \rot{CIFAR100} && \rot{Birdsnap} && \rot{SUN397} &&  \rot{Cars} &&  \rot{Aircraft}  && \rot{VOC2007} && \rot{DTD}  &&  \rot{Pets} &&  \rot{Caltech-101} &&  \rot{Flowers} && \rot{\textbf{Average}} \\

    \midrule
    \multirow{8}{=}{\centering \rotatebox{90}{YFCC}}
    && BYOL-1k          && 67.9	&& 85.0 && 63.9 && 31.3 && 63.4 && 44.3 && 47.5 && 81.8 && 75.2 && 71.1 && 84.0 && 93.4 && 67.4 \\
    && BYOL-3k          && 68.8	&& 86.5 && 66.6 && 33.2 && 63.9 && 46.5 && 49.8 && 82.3 && 73.6 && 73.9 && 86.5 && 95.4 && 68.9 \\
    && BYOL-5k          && 69.1	&& 85.8	&& 66.8	&& \textbf{35.5}	&& 64.1	&& 50.1	&& \textbf{51.9}	&& 82.5	&& 74.5	&& 74.0	&& \textbf{87.6}	&& 95.8	&& 69.8 \\
    && \baseline{}-1k   && 67.7	&& 87.8	&& 69.9	&& 29.4	&& 63.4	&& 41.1	&& 45.6	&& 81.6	&& 75.8	&& 67.7	&& 85.6	&& 92.9	&& 67.4 \\
    && \baseline{}-3k   && 67.9	&& \textbf{88.3}	&& 70.2	&& 29.6	&& 63.8	&& 40.7	&& 45.9	&& 82.4	&& 76.0	&& 69.2	&& 85.4	&& 92.3	&& 67.6 \\
    && \baseline{}-5k   && 68.4	&& 87.6	&& 69.7	&& 30.5	&& 63.9	&& 41.0	&& 46.7	&& 82.4	&& 76.2	&& 68.5	&& 86.0	&& 93.0	&& 67.8 \\
    && DnC-3k  && 71.9	&& 87.3	&& 70.1	&& 34.4	&& 65.7	&& 48.2	&& 46.3	&& 82.7	&& 75.5	&& 75.7	&& 86.0	&& 96.5	&& \cellcolor{DnCBG}70.0 \\
    && DnC-4.5k           && \textbf{72.1} && \textbf{88.0} && \textbf{71.1} && \textbf{35.5} && \textbf{67.2} && \textbf{52.6} && 49.2 && \textbf{83.7} && \textbf{76.5} && \textbf{75.9} && 87.0 && \textbf{97.8} && \cellcolor{DnCBG}\textbf{71.4} \\

    \midrule
    \multirow{8}{=}{\centering \rotatebox{90}{\jft{}}}
    && BYOL-1k          && 72.7 && 90.1 && 71.7 && 33.9 && 61.0 && 62.4 && 52.1 && 81.1 && 74.9 && 76.0 && 89.0 && 94.3 && 71.6 \\
    && BYOL-3k          && 72.8 && 89.9 && 72.5 && 36,7 && 62.1 && 63.3 && 53.2 && 81.6 && 75.5 && 77.8 && 89.5 && 94.5 && 72.5 \\
    && BYOL-5k          && 73.3 && 89.8 && 72.4 && 38.2 && 61.8 && 64.4 && \textbf{54.4} && 81.3 && 75.5 && 77.0 && \textbf{90.1} && 94.3 && 72.7 \\
    && \baseline{}-1k   && 71.9 && 90.4 && 72.7 && 32.8 && 61.3 && 59.3 && 51.6 && 81.5 && 75.4 && 74.5 && 89.3 && 93.9 && 71.2 \\
    && \baseline{}-3k   && 72.7 && 90.8 && 73.0 && 33.5 && 62.2 && 59.8 && 51.6 && 81.4 && \textbf{77.3} && 76.2 && 88.7 && 93.5 && 71.7 \\
    && \baseline{}-5k   && 72.8 && 90.7 && 72.5 && 33.8 && 62.2 && 60.6 && 50.9 && 81.9 && 75.3 && 75.8 && 89.5 && 93.8 && 71.7  \\
    && DnC-3k   && 74.8 && \textbf{91.6} && \textbf{74.9} && 38.2 && 63.8 && 68.6 && 53.4 && \textbf{83.0} && \textbf{77.1} && 82.5 && \textbf{90.5} && 97.2 && \cellcolor{DnCBG}74.6  \\
    && DnC-4.5k      && \textbf{78.7} && \textbf{91.7} && \textbf{74.9} && \textbf{42.1} && \textbf{65.0} && \textbf{75.3} && \textbf{54.1} && \textbf{83.1} && 76.6 && \textbf{86.1} && \textbf{90.2} && \textbf{98.2} && \cellcolor{DnCBG}\textbf{76.3} \\

    \bottomrule
  \end{tabularx}
\end{table*}

%% file: fig_tex/table_finetune_complete.tex
\begin{table*}[th]
  \setlength{\tabcolsep}{5.8pt}
  \setlength{\extrarowheight}{5pt}
  \renewcommand{\arraystretch}{0.75}
  \centering
  \small
  \caption{Fine-tuning pre-trained model for transfer learning experiments, including object detection on COCO dataset, semantic segmentation on Pascal VOC 2012, and depth estimation on NYU v2 dataset. For the evaluation metrics of \emph{rms} and \emph{rel}, lower is better.}
  \label{tbl:transfer_finetune_complete}
  
  \begin{tabular}{ll|cccccc|c|ccccc}
  \toprule
  & & \multicolumn{6}{c|}{COCO object detection, 1x schedule} & Seg. & \multicolumn{5}{c}{NYU v2 depth estimation}\\
  
  & & \apbbox{~} & \apbbox{50} & \apbbox{75} & \apmask{~} & \apmask{50} & \apmask{75} & mIoU & $<$1.25 & $<$1.25$^2$ & $<$1.25$^3$ & rms$\mathcolor{red}{\downarrow}$ & rel$\mathcolor{red}{\downarrow}$ \\
  \midrule
  
  \multicolumn{2}{c|}{ImageNet Super.} & 39.5 & 60.1 & 43.3 & 35.4 & 56.9 & 38.1 & 74.4 & 81.1 & 95.3 & 98.8 & 0.573 & 0.127 \\
  
  \midrule

  \multirow{3}{*}{\small \rotatebox{90}{ImageNet}}
  & BYOL-3k & 40.9\cgaphl{+}{1.4} & 61.9 & 45.0 & 36.7\cgaphl{+}{1.3} & 58.5 & 39.2 & 76.3 & 84.7 & 97.0 & 99.1 & 0.525 & 0.126 \\
  & \baseline-3k & 41.5\cgaphl{+}{2.0} & 62.3 & 45.4 & 37.0\cgaphl{+}{1.6} & 59.0 & 39.7 & 76.2 & 84.6 & 97.0 & 99.3 & 0.527 & 0.126 \\
  & DnC-3k & \cellcolor{DnCBG}41.7\cgaphl{+}{2.2}	   & \cellcolor{DnCBG}62.6 & \cellcolor{DnCBG}45.6 & \cellcolor{DnCBG}37.3\cgaphl{+}{1.9} & \cellcolor{DnCBG}59.2 & \cellcolor{DnCBG}40.1 & \cellcolor{DnCBG}76.9 & \cellcolor{DnCBG}85.1 & \cellcolor{DnCBG}97.0 & \cellcolor{DnCBG}99.2 & \cellcolor{DnCBG}0.525 & \cellcolor{DnCBG}0.124 \\

  \midrule
  
  \multirow{8}{*}{\small \rotatebox{90}{YFCC}}
  & BYOL-1k & 40.8\cgaphl{+}{1.3} & 61.9 & 45.0 & 36.4\cgaphl{+}{1.0} & 58.4 & 38.8 & 75.5 & 85.8	& 97.2 & 99.2	& 0.511 & 0.122 \\
  & BYOL-3k & 41.0\cgaphl{+}{1.5} & 61.6 & 45.0 & 36.6\cgaphl{+}{1.2} & 58.5 & 39.2 & 75.5 & 85.2 & 96.9 & 99.0
  & 0.537 & 0.124 \\
  & BYOL-5k & 41.1\cgaphl{+}{1.6} & 62.0 & 45.1 & 36.6\cgaphl{+}{1.2} & 58.6 & 38.9 & 75.1 & 83.5 & 96.4 & 99.0 & 0.558 & 0.130 \\
  
  & \baseline-1k & 40.2\cgaphl{+}{0.7} & 61.1 & 44.2 & 36.0\cgaphl{+}{0.6} & 57.8& 38.2 & 75.0 & 85.7 & 97.1 & 99.3 & 0.515 & 0.122 \\
  & \baseline-3k & 40.7\cgaphl{+}{1.2} & 61.6 & 44.4 & 36.3\cgaphl{+}{0.9} & 58.3& 38.8 & 75.3 & 86.6 & 97.2 & 99.3 &	0.502 & 0.120 \\
  & \baseline-5k & 40.8\cgaphl{+}{1.3} & 61.7 & 44.8 & 36.6\cgaphl{+}{1.2} & 58.5& 39.0 & 75.5 & 86.7 & 97.4 & 99.3 & 0.503 & 0.117 \\
  
  & DnC-3k & \cellcolor{DnCBG}41.0\cgaphl{+}{1.5} & \cellcolor{DnCBG}61.6 & \cellcolor{DnCBG}44.7 & \cellcolor{DnCBG}36.6\cgaphl{+}{1.2} & \cellcolor{DnCBG}58.5 & \cellcolor{DnCBG}39.5 & \cellcolor{DnCBG}76.1 & \cellcolor{DnCBG}86.7 & \cellcolor{DnCBG}97.3 & \cellcolor{DnCBG}99.3 & \cellcolor{DnCBG}0.506 & \cellcolor{DnCBG}0.117\\
  & DnC-4.5k & \cellcolor{DnCBG}41.5\cgaphl{+}{2.0} & \cellcolor{DnCBG}62.5 & \cellcolor{DnCBG}45.6 & \cellcolor{DnCBG}37.0\cgaphl{+}{1.6} & \cellcolor{DnCBG}59.3 & \cellcolor{DnCBG}39.6 & \cellcolor{DnCBG}76.6 & \cellcolor{DnCBG}86.2 & \cellcolor{DnCBG}97.2 & \cellcolor{DnCBG}99.3 & \cellcolor{DnCBG}0.512 & \cellcolor{DnCBG}0.121\\
  
  \midrule
  \multirow{8}{*}{\small \rotatebox{90}{\jft{}}}
  & BYOL-1k & 40.5\cgaphl{+}{1.0} & 61.3 & 44.4 & 36.4\cgaphl{+}{1.0} & 58.2 & 38.8 & 75.5 & 85.8 & 97.1 & 99.2 &	0.519 & 0.121 \\
  & BYOL-3k & 40.5\cgaphl{+}{1.0} & 61.1 & 44.7 & 36.4\cgaphl{+}{1.0} & 57.9 & 39.2 & 75.7 & 85.6 & 97.0 & 99.2 & 0.525 & 0.122 \\
  & BYOL-5k & 40.6\cgaphl{+}{1.1} & 61.2 & 44.3 & 36.2\cgaphl{+}{0.8} & 58.1 & 38.8 & 75.8 & 84.4 & 96.5 & 99.0 & 0.544 & 0.129 \\
  
  & \baseline-1k  & 40.3\cgaphl{+}{0.8} & 61.0 & 44.2 & 36.3\cgaphl{+}{0.9} & 58.0 & 38.8 & 75.7 & 84.9 & 96.8 & 99.2 & 0.526 & 0.126\\
  & \baseline-3k  & 40.5\cgaphl{+}{1.0} & 61.2 & 44.4 & 36.4\cgaphl{+}{1.0} & 58.1 & 39.0 & 75.8 & 85.9 & 97.2& 99.3 & 0.514 & 0.121\\
  & \baseline-5k  & 41.1\cgaphl{+}{1.6} & 62.0 & 45.4 & 36.9\cgaphl{+}{1.5} & 58.9 & 39.5 & 76.1 & 86.3 & 97.2 & 99.3 & 0.513 & 0.120\\
  
  & DnC-3k & \cellcolor{DnCBG}41.6\cgaphl{+}{2.1} & \cellcolor{DnCBG}62.3 & \cellcolor{DnCBG}45.5 & \cellcolor{DnCBG}37.2\cgaphl{+}{1.8} & \cellcolor{DnCBG}59.1 & \cellcolor{DnCBG}39.8 & \cellcolor{DnCBG}76.8 & \cellcolor{DnCBG}86.0 & \cellcolor{DnCBG}97.3 & \cellcolor{DnCBG}99.3 & \cellcolor{DnCBG}0.517 & \cellcolor{DnCBG}0.119\\
  & DnC-4.5k & \cellcolor{DnCBG}41.7\cgaphl{+}{2.2} & \cellcolor{DnCBG}62.5 & \cellcolor{DnCBG}45.9 & \cellcolor{DnCBG}37.2\cgaphl{+}{1.8} & \cellcolor{DnCBG}59.3 & \cellcolor{DnCBG}39.8 & \cellcolor{DnCBG}76.9 & \cellcolor{DnCBG}86.1 & \cellcolor{DnCBG}97.2 & \cellcolor{DnCBG}99.4 & \cellcolor{DnCBG}0.509 & \cellcolor{DnCBG}0.119\\
  
  \bottomrule
  \end{tabular}
\vspace{-4mm}
\end{table*}